\documentclass[letterpaper]{article} 
\usepackage{aaai2027}  
\nocopyright
\usepackage[hyphens]{url}  
\usepackage{graphicx} 
\usepackage{natbib}  
\usepackage{caption} 
\usepackage{algorithm}
\usepackage{algorithmic}
\usepackage{amsmath}
\usepackage{amssymb}
\usepackage{amsfonts}
\usepackage{graphicx}
\usepackage{tabularx}
\usepackage{amsthm} 

\usepackage{array}
\usepackage{algorithm}
 \usepackage{booktabs}
 \usepackage{color}
\usepackage[table]{xcolor}  
 \usepackage{multirow} 
\usepackage{newfloat}
\usepackage{listings}
\DeclareCaptionStyle{ruled}{labelfont=normalfont,labelsep=colon,strut=off} 
\floatstyle{ruled}
\newfloat{listing}{tb}{lst}{}
\floatname{listing}{Listing}

\usepackage{booktabs}

\title{Gaussian Core LoRA: Distribution-Aware Dynamic Adaptation for Broad Concept Erasure}
\author {
    Qinghui Gong\textsuperscript{\rm 1}
    Xunlei Chen\textsuperscript{\rm 2},
    Yu-Xuan Zhang\textsuperscript{\rm 1},
    Hua Meng\textsuperscript{\rm 1},
    Zhengchun Zhou\textsuperscript{\rm 1}\
}

\affiliations {
    \textsuperscript{\rm 1}Southwest Jiaotong University\quad
    \textsuperscript{\rm 2}University of Electronic Science and Technology of China
}

\begin{document}

\maketitle

\begin{abstract}
Concept erasure aims to suppress unsafe, privacy-sensitive, or undesirable
generations in text-to-image diffusion models while preserving benign
semantics, visual quality, and deployment efficiency. Existing adapter-based
methods, such as Low-Rank Adaptation (LoRA), typically freeze the diffusion
backbone and learn lightweight parameter updates to steer generation away
from target semantics. However, these methods usually assign a static semantic
erasure direction to each target concept. This assumption is overly coarse for broad and
complex target concepts, since a concept often contains multiple latent
semantic prototypes involving different objects, scenes, or relations, and
requires different local erasure directions. A single LoRA update averages
these heterogeneous erasure demands, leading to under-erasure on difficult
prototypes and over-editing of nearby benign semantics.
To address this limitation, we propose Gaussian Core LoRA, a
distribution-aware low-rank adaptation framework. It fits a Gaussian mixture
model in the prompt feature space to estimate latent semantic prototypes
within the target concept. During inference, each input prompt is projected
into this feature space to compute its Gaussian posterior responsibilities,
which condition the core generator to produce a prompt-specific,
norm-bounded residual reconfiguration of the shared LoRA rank space. This
enables prototype-adaptive erasure with a single lightweight adapter. Compared with
the strongest baseline on each metric, Gaussian Core LoRA reduces average
Attack Success Rate (ASR) by 7.95\%, lowers COCO Fr\'echet Inception Distance
(FID) by 14.72\%, and improves CLIP Score by 4.98\%. Further experiments show
robustness to adversarial prompts, scalability to multi-identity and
multi-style erasure, and compatibility with SDXL and FLUX.
\end{abstract}

\begin{figure}[t]
    \centering
    \includegraphics[width=\linewidth]{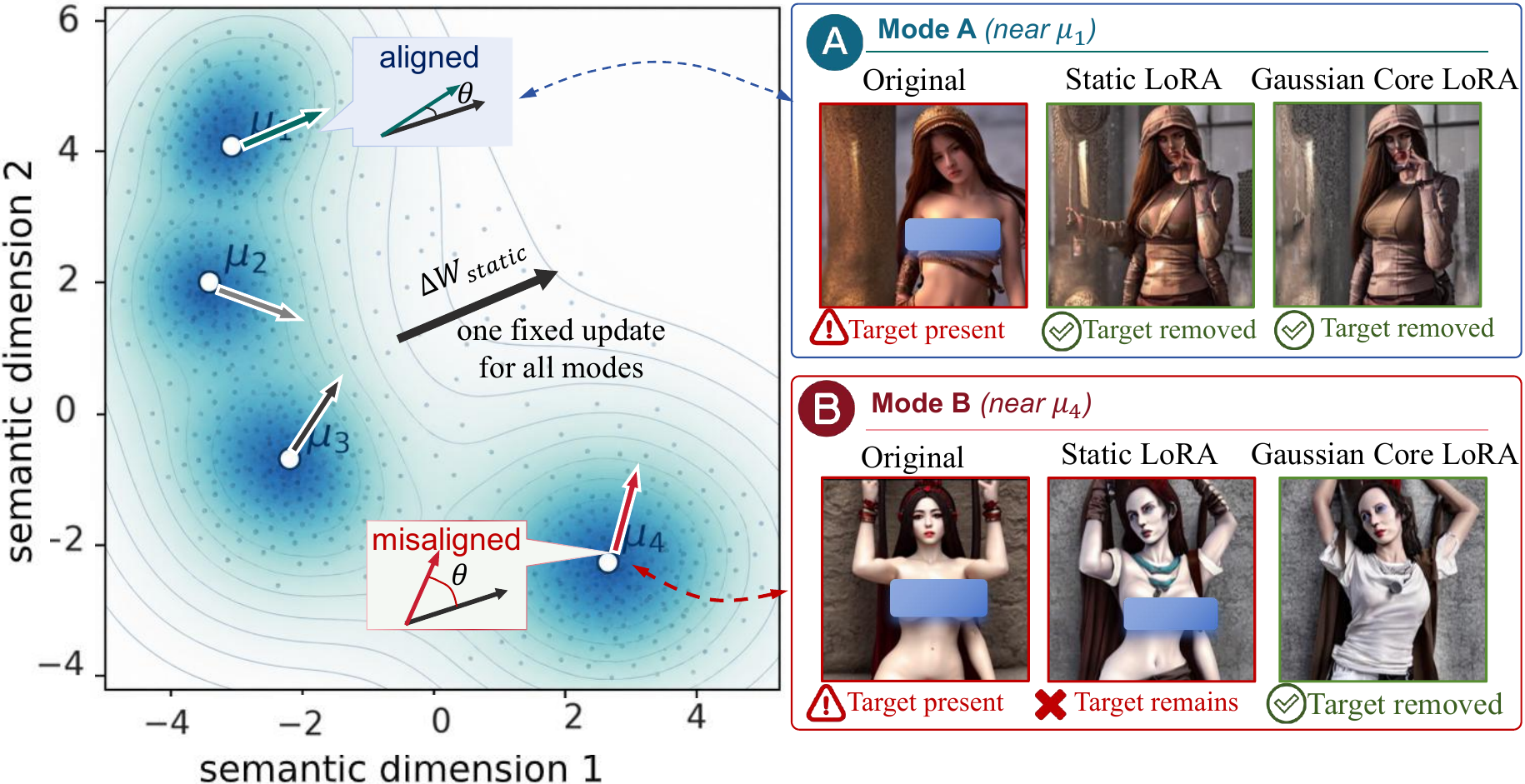}
  \caption{
Motivation of Gaussian Core LoRA.
An unsafe concept may contain multiple latent modes requiring different erasure directions.
Static LoRA uses a fixed update and may fail on misaligned modes, whereas Gaussian Core LoRA adapts the update to each prompt's latent-mode position.
}
    \label{fig:motivation}
\end{figure}

\section{Introduction}
Recent advances in text-to-image diffusion models have substantially improved open-ended visual generation~\cite{song2020denoising,tu2025driveditfit,ho2020denoising}. However, their large-scale training data inevitably contains undesirable concepts, such as sexual, violence, hate, sensitive identities, or copyrighted characters~\cite{jiang2023ai,wang2025precise,lu2024mace}, exposing deployed models to safety, privacy, and ethical risks. Therefore, how to suppress target concepts without retraining the entire model, while preserving benign semantics, visual quality, and deployment efficiency, has become an important problem in diffusion model safety control.

Existing concept erasure methods can be roughly categorized into inference-time intervention, model fine-tuning, and parameter-efficient adaptation. Inference-time methods require no training~\cite{liu2024latent,wu2024universal,yoon2025safree,wang2025precise} and are easy to deploy, but the model's internal generative tendency is not truly modified, making them vulnerable to implicit prompts, prompt rewriting, or adversarial attacks~\cite{zhang2024generate,chin2023prompting4debugging,tsai2024ring}. Global or local fine-tuning methods can more directly change model behavior~\cite{gandikota2023erasing,gandikota2024unified,feng2025fg,tu2025sdwv,fan2024salun}, but they incur high training costs and may damage non-target semantics. In contrast, Low-Rank Adaptation (LoRA)-based adapters freeze the diffusion backbone and perform concept erasure with only a small number of pluggable parameters~\cite{kumari2023ablating,zhang2024unlearncanvas,hu2022lora}, providing a practical trade-off among training cost, storage overhead, and deployment flexibility.
However, most LoRA-style erasure methods still follow concept-level static modeling: given a human-defined target concept, they learn a fixed low-rank update and apply it to all prompts under that concept. Early methods train independent adapters for individual persons, styles, objects, or sensitive words~\cite{zhao2024separable,lyu2024one,liu2025dyme}, while recent multi-concept methods, such as MACE~\cite{lu2024mace} and SuPLoRA~\cite{tu2026mass}, rely on adapter fusion or shared adaptation to improve scalability. Although these methods exploit shared structures among related concepts, they are still organized around manually defined concept boundaries rather than the actual distribution of target prompts in the model feature space.

This label-level formulation overlooks the internal diversity of how diffusion models represent a target concept. A human label only specifies what should not be generated, but not how the model represents and composes that concept. A broad concept may contain multiple identity instances, visual modes, scene relations, or semantic subcategories, which occupy different regions in the prompt feature space and require different local erasure directions~\cite{cassano2025saemnesia,cai2026prototype,yin2025rethinking}. As illustrated in Fig.~\ref{fig:motivation}, even a manually defined unsafe concept can form separated latent clusters, where a fixed LoRA correction may cover only part of the modes. As a result, static LoRA compresses a high-entropy target distribution into a single fixed update, causing mode averaging: weak updates fail on difficult modes, while strong updates may interfere with nearby benign semantics~\cite{jiang2026himole}.
This motivates a new perspective: \textbf{\textit{broad concept erasure should be viewed
as learning a conditional low-rank editing operator over the prompt
feature distribution, rather than estimating a single parameter update
for a human-defined label.}}

Based on this view, we propose Gaussian Core LoRA, a distribution-aware
dynamic low-rank adaptation method for broad concept erasure. It learns
a shared LoRA space for a broad target concept and dynamically
reconfigures this space according to each prompt's position in the
feature distribution. Specifically, a Gaussian support gate suppresses adapter activation for prompts far from the target distribution, GMM responsibilities provide soft latent-mode routing, and a dynamic core produces prompt-conditioned rank-space reconfiguration. Empirically, Gaussian Core LoRA improves the
erasure--preservation trade-off across seven unsafe prompt groups,
reducing average ASR by \(7.95\%\) relative to the strongest average
baseline while improving COCO FID and CLIP Score. Its adaptive
rank-space reconfiguration also improves robustness to
adversarial prompts, achieving \(2.0\%\), \(14.5\%\), and
\(10.0\%\) ASR on Ring-A-Bell~\cite{tsai2024ring}, P4D~\cite{chin2023prompting4debugging}, and UnDiff~\cite{zhang2024generate}, respectively, while
remaining effective on both SDXL~\cite{podell2024sdxl} and FLUX~\cite{esser2024scaling}.

Our main contributions are summarized as follows:
\begin{itemize}
\item We formulate concept erasure from a feature-distribution perspective, showing that multimodal concept distributions make static LoRA prone to mode averaging, incomplete mode coverage, and semantic interference.

\item We propose Gaussian Core LoRA, a gate-router-core dynamic low-rank adaptation framework. It uses a Gaussian support gate for coarse activation screening, GMM
posterior responsibilities for soft latent-mode routing, and a dynamic
core for prompt-conditioned reconfiguration in the shared LoRA rank
space.

\item We validate that Gaussian Core LoRA uses a single shared adapter
to cover heterogeneous modes within each broad target concept without
separate mode-specific training. The same design scales to joint
erasure of up to 50 identities/styles and maintains favorable
erasure--preservation trade-offs under adversarial prompts and
cross-architecture evaluation on SDXL and FLUX.

\end{itemize}

\section{Related Work}

\subsection{Inference-Time Intervention and Model Tuning}

Concept erasure methods for text-to-image diffusion models mainly include inference-time intervention and model tuning. Inference-time methods suppress undesired concepts by modifying denoising guidance, prompts, conditioning signals, attention responses, or semantic subspaces without updating model parameters~\cite{schramowski2023safe,wu2024universal,yoon2025safree,wang2025precise}. Although efficient, they leave the underlying model unchanged and can be bypassed by prompt rewriting, module disabling, or adversarial prompts~\cite{rando2022red}. Model tuning methods directly update model parameters to remove objects, styles, identities, or unsafe content~\cite{gandikota2023erasing,heng2023selective,kumari2023ablating,fan2024salun}, and have been extended to multi-concept or large-scale erasure~\cite{gandikota2024unified,zhao2024separable}. These methods provide stronger behavioral modification but often incur higher training costs and may disturb non-target semantics.

\subsection{Adapter-Based Concept Erasure}

Parameter-efficient adaptation offers a practical trade-off by freezing the diffusion backbone and learning lightweight modules~\cite{hu2022lora,kumari2023ablating}. Recent adapter-based methods improve scalability by training, merging, sharing, or dynamically composing LoRA-style modules. MACE merges concept-specific LoRA modules~\cite{lu2024mace}, while other methods explore separable, shared, dynamic, or hierarchical adapters to reduce interference across concepts~\cite{lyu2024one,liu2025dyme,tu2026mass}. These methods show that related concepts may share semantic directions, attention responses, or low-rank erasure structures.

However, most adapter-based methods still rely on predefined concept units and fixed adapter updates. For broad concepts with heterogeneous latent modes, this static formulation may average different editing demands. This motivates a distribution-aware perspective, where the erasure update should be dynamically adjusted according to the input prompt rather than solely determined by a human-defined concept label.

\begin{figure*}[t]
\centering
\includegraphics[width=0.9\textwidth]{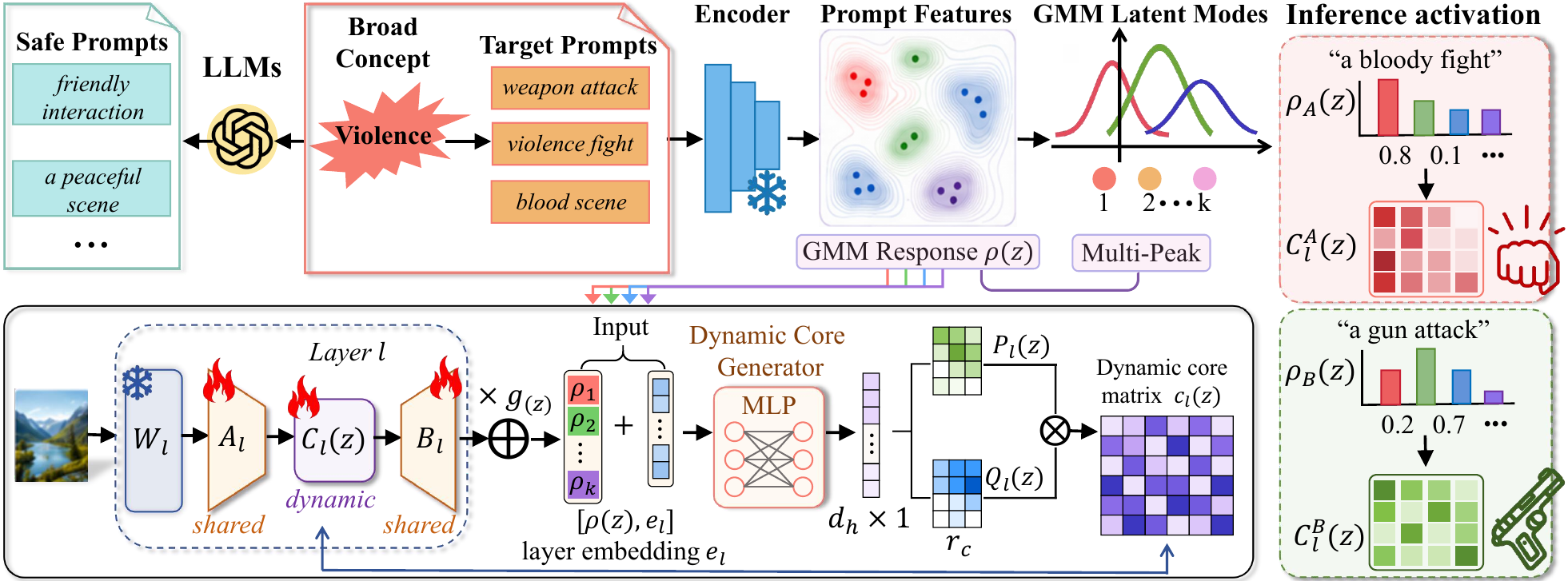}
\caption{
Overview of Gaussian Core LoRA.
Sensitive prompts and safe semantics are first constructed from a target concept.
Prompt features are modeled by a GMM to obtain latent mode responsibilities and Gaussian support regions.
The gate \(g(z)\) determines whether to activate the adapter branch, while the responsibility vector \(\rho(z)\) determines how to reconfigure the shared LoRA rank space.
Safe semantics provide paired erasure and retention constraints during training.
During inference, each prompt is projected into the same feature space to generate layer-aware dynamic cores \(C_l(z)\), producing prompt-dependent erasure directions across diffusion layers.
}
\label{fig:overflow}
\end{figure*}

\section{Method}

\subsection{Overview}
We formulate broad concept erasure as a distribution-aware
parameter-efficient adaptation problem. Given a pretrained
text-to-image diffusion model \(\epsilon_{\theta}\) and a broad concept
\(c\), target prompts are treated as samples from an underlying
prompt-feature distribution. The goal is to learn a lightweight adapter
\(\Delta_{\psi}\) that suppresses target prompts while preserving
non-target and semantically neighboring benign prompts.

As shown in Fig.~\ref{fig:overflow}, Gaussian Core LoRA follows a
gate-router-core design. The Gaussian support gate provides coarse activation screening, GMM
responsibilities perform soft mode routing, and a dynamic core
reconfigures the shared LoRA rank space. For
the \(l\)-th LoRA-injected layer, the effective update is
\begin{equation}
\Delta W_l(z)=g(z)B_l C_l(z) A_l ,
\label{eq:gclora_update}
\end{equation}
where \(z\) is the prompt feature, \(g(z)\in[0,1]\) is the Gaussian
support gate, \(A_l\) and \(B_l\) are shared LoRA bases, and \(C_l(z)\) is a
prompt-conditioned core matrix. This design reuses shared erasure
structures while producing mode-adaptive updates.

\subsection{Task-Feature Space and Mode Modeling}

Given a broad concept \(c\), we construct a target prompt set
\begin{equation}
\mathcal{D}_{t}=\{p_i^{-}\}_{i=1}^{N_t},
\end{equation}
where \(p_i^{-}\) is the \(i\)-th prompt containing the concept to be erased, and \(N_t\) is the number of target prompts.

For each prompt \(p\), we extract token-level representations with the frozen text encoder:
\begin{equation}
E(p)=[e_1,e_2,\ldots,e_T],
\end{equation}
where \(e_t\) is the feature of the \(t\)-th token.
We then pool valid token features to obtain a prompt-level representation:
\begin{equation}
h(p)=\frac{1}{|\mathcal{M}|}\sum_{t\in\mathcal{M}}e_t ,
\end{equation}
where \(\mathcal{M}\) is the set of valid token indices.

To obtain a compact task representation, we project the prompt-level
feature into a low-dimensional space:
\begin{equation}
z=P^{\top}(h(p)-\bar{h}),
\end{equation}
where \(P\) is the PCA projection matrix and \(\bar{h}\) is the mean
feature of target prompts. 
Then we fit a diagonal-covariance GMM on the target task features \(\{z_i\}_{i=1}^{N_t}\):
\begin{equation}
p_t(z)=p(z|c)=\sum_{k=1}^{K}\pi_k \mathcal{N}(z;\mu_k,\Sigma_k),
\label{eq:gmm}
\end{equation}
where \(K\) is the number of latent modes, \(\pi_k\) is the mixture weight, \(\mu_k\) is the mean, and \(\Sigma_k\) is the diagonal covariance of the \(k\)-th component:
\begin{equation}
\Sigma_k=\mathrm{diag}(\sigma_{k,1}^{2},\ldots,\sigma_{k,d}^{2}).
\end{equation}
The fixed GMM parameters provide a reference distribution for
target-mode routing.

For a prompt feature \(z\), its posterior responsibility for the \(k\)-th target mode is
\begin{equation}
\rho_k(z)=
\frac{\pi_k \mathcal{N}(z;\mu_k,\Sigma_k)}
{\sum_{j=1}^{K}\pi_j \mathcal{N}(z;\mu_j,\Sigma_j)} .
\label{eq:responsibility}
\end{equation}
The responsibility vector is \(\rho(z)=[\rho_1(z),\ldots,\rho_K(z)]\), with \(\sum_{k=1}^{K}\rho_k(z)=1\).
It describes the soft position of the prompt among target latent modes.

\subsection{Gaussian Support Gate}

The responsibility vector \(\rho(z)\) only describes the relative
assignment of a prompt among target latent modes. Since it is
normalized over the target Gaussian components, a benign or open-set
prompt can still receive a confident mode assignment, causing unnecessary adapter activation on non-target content.

To address this issue, we introduce a Gaussian support gate derived
from the fitted target GMM. For each target mode, we compute the
Mahalanobis distance between the prompt feature \(z\) and the Gaussian
center:
\begin{equation}
d_k(z)=
(z-\mu_k)^\top\Sigma_k^{-1}(z-\mu_k).
\end{equation}
This distance measures whether \(z\) falls within the elliptical
support of the \(k\)-th target mode.

For each mode, we estimate its support radius from target training
prompts:
\begin{equation}
\tau_k=
Q_{\beta}^{w}
\left(
\{d_k(z_i)\}_{i=1}^{N_t},
\{\rho_k(z_i)\}_{i=1}^{N_t}
\right),
\end{equation}
where \(Q_{\beta}^{w}\) denotes the responsibility-weighted
\(\beta\)-percentile, and the selection of \(\beta\) is detailed in
App.~\ref{app:details}. The normalized distance to the target
multi-mode support is then defined as
\begin{equation}
r(z)=
\min_k
\frac{d_k(z)}{\tau_k}.
\end{equation}
A prompt is considered inside the target concept support if it is close
to at least one Gaussian mode, i.e., \(r(z)\leq 1\).

The final gate is computed as
\begin{equation}
g(z)=
\sigma\left(
\frac{1-r(z)}{T_g}
\right),
\end{equation}
where \(T_g=0.2\) controls the boundary softness. The gate strengthens
the adapter update inside the target Gaussian support and suppresses it
outside all target supports.

\subsection{Gaussian Core LoRA}
We inject Gaussian Core LoRA into the query, key, value, and output projection layers of the cross-attention blocks. For the
\(l\)-th injected projection layer, let the frozen original weight be \(W_l\in\mathbb{R}^{d_{\mathrm{out}}\times d_{\mathrm{in}}}\).
A standard LoRA update is
\begin{equation}
\Delta W_l=B_lA_l,
\end{equation}
where \(A_l\in\mathbb{R}^{r\times d_{\mathrm{in}}}\),
\(B_l\in\mathbb{R}^{d_{\mathrm{out}}\times r}\), and \(r\) is the LoRA rank. 

Gaussian Core LoRA gates the adapter update and inserts a dynamic core into the LoRA rank space:
\begin{equation}
\Delta W_l(z)=g(z)B_l C_l(z) A_l,
\end{equation}
where \(g(z)\in[0,1]\) controls the intervention strength and
\(C_l(z)\in\mathbb{R}^{r\times r}\) is a prompt-conditioned core matrix.
Given the layer input \(u_l\), the layer output is $y_l=W_lu_l+\Delta W_l(z)u_l,$ \(A_l\) and \(B_l\) define a shared erasure subspace, while
\(C_l(z)\) reconfigures this subspace according to the target-mode
responsibility \(\rho(z)\).

\begin{table*}[t]
\centering
\caption{
Broad unsafe-concept erasure on seven I2P prompt groups. We report
Q16-based ASR (\%) for erasure effectiveness, and COCO FID and CLIP
Score for general-content preservation, averaged across the seven
category-specific edited models.
}
\label{tab:unsafe_erasure}
\small
\setlength{\tabcolsep}{4.0pt}
\begin{tabular}{lccccccccc}
\toprule
Method
& Hate$\downarrow$
& Harassment$\downarrow$
& Illegal Activity$\downarrow$
& Self-harm$\downarrow$
& Sexual$\downarrow$
& Shocking$\downarrow$
& Violence$\downarrow$
& COCO FID$\downarrow$
& COCO CLIP$\uparrow$ \\
\midrule

SD v1.5
& 21.5\%
& 19.3\%
& 19.8\%
& 35.0\%
& 55.0\%
& 41.8\%
& 39.5\%
& -
& 28.6 \\

ESD-x
& \underline{3.8\%}
& 6.8\%
& 16.3\%
& 11.0\%
& 16.0\%
& 15.8\%
& 6.8\%
& 36.8
& 25.0 \\

RECE
& 4.5\%
& 6.3\%
& 6.3\%
& 8.8\%
& 8.3\%
& 9.5\%
& 13.8\%
& 39.2
& 24.6 \\

MACE
& 4.0\%
& \underline{6.0\%}
& 6.0\%
& 6.0\%
& 3.0\%
& 10.0\%
& 8.0\%
& 38.5& 23.8 \\

SAFREE
& 5.0\%
& 9.5\%
& 11.0\%
& 7.0\%
& 5.5\%
& 13.5\%
& 9.5\%
& 34.6
& \underline{26.1} \\

AdaVD
& 10.0\%
& 8.0\%
& 8.5\%
& 10.8\%
& 2.8\%
& 17.0\%
& 12.5\%
& 43.8& 23.3 \\

Proto-Guided
& \textbf{3.5\%}
& \textbf{5.8\%}
& \underline{5.8\%}
& \underline{4.5\%}
& \underline{1.8\%}
& \textbf{7.5\%}
& \underline{6.3\%}
& \underline{26.5}& 25.7 \\

Ours
& 4.0\%
& 6.3\%
& \textbf{5.0\%}
& \textbf{3.8\%}
& \textbf{0.0\%}
& \underline{7.8\%}
& \textbf{5.5\%}
& \textbf{22.6}& \textbf{27.4} \\

\bottomrule
\end{tabular}
\end{table*}

To make the core layer-aware, each injected layer is assigned a learnable embedding \(e_l\in\mathbb{R}^{d_e}\).
The core generator takes the target-mode responsibility and layer embedding as input: $q_l(z)=[\rho(z);e_l],$ where \([\cdot;\cdot]\) denotes concatenation.
A shared generator \(G_{\phi}\) outputs two low-rank core factors:
\begin{equation}
[P_l(z),Q_l(z)]=G_{\phi}(q_l(z)),
\end{equation}
where \(P_l(z),Q_l(z)\in\mathbb{R}^{r\times r_c}\), and \(r_c\) is the residual core rank.
We then define the unnormalized residual core as
\begin{equation}
R_l(z)=\tanh(P_l(z))\tanh(Q_l(z))^{\top}.
\end{equation}
To explicitly control the core perturbation scale, we normalize it by its Frobenius norm:
\begin{equation}
\widehat{R}_l(z)=
\frac{R_l(z)}{\|R_l(z)\|_F+\varepsilon}.
\end{equation}
The dynamic core is then defined as
\begin{equation}
C_l(z)=I_r+\eta\widehat{R}_l(z),
\label{eq:core}
\end{equation}
where \(\eta\) controls the maximum core reconfiguration strength, and \(r_c\) controls the rank capacity of the residual core.
This normalized identity-residual form gives
\begin{equation}
\|C_l(z)-I_r\|_F
=
\eta
\frac{\|R_l(z)\|_F}{\|R_l(z)\|_F+\varepsilon}
\leq \eta.
\end{equation}
The gate bounds the effective residual reconfiguration:
\begin{equation}
\|g(z)(C_l(z)-I_r)\|_F
\leq g(z)\eta.
\end{equation}
Therefore, the gate controls the activation strength of the whole adapter, while the dynamic core provides a bounded prompt-conditioned reconfiguration inside the shared LoRA rank space.
Details are provided in Appendix~\ref{app:matrix_perturbation}.

\subsection{Joint Training Objective}

We train Gaussian Core LoRA following the standard LoRA-based concept erasure paradigm.
For each target prompt \(p_i^{-}\), we construct a safe replacement prompt \(p_i^{+}\) with a large language model:
\begin{equation}
p_i^{+}=\mathrm{LLM}(p_i^{-},c).
\end{equation}
The replacement prompt removes or substitutes the target concept \(c\), while preserving the main non-target context.
This gives a text-pair set
\begin{equation}
\mathcal{P}=\{(p_i^{-},p_i^{+})\}_{i=1}^{N_t}.
\end{equation}

For a text pair \((p_i^{-},p_i^{+})\), we align the denoising prediction of the modified model under the target prompt with the prediction of the frozen model under the safe prompt:
\begin{equation}
\mathcal{L}_{\mathrm{pair}}
=
\mathbb{E}_{(p_i^{-},p_i^{+})\in\mathcal{P},t}
\left[
\left\|
\epsilon_{\theta+\Delta_{\psi}}(x_t,t,p_i^{-})
-
\epsilon_{\theta}(x_t,t,p_i^{+})
\right\|_2^2
\right],
\end{equation}
where \(x_t\) is the noisy latent at timestep \(t\), \(\epsilon_{\theta+\Delta_{\psi}}\) is the model equipped with Gaussian Core LoRA, and \(\epsilon_{\theta}\) is the frozen original model.

Since \(p_i^+\) removes the target concept while preserving the main
non-target context, it also serves as a natural retention prompt. We
therefore constrain the edited model to match the frozen model on
\(p_i^+\):
\begin{equation}
\mathcal{L}_{\mathrm{retain}}
=
\mathbb{E}_{p_i^+,t}
\left[
\left\|
\epsilon_{\theta+\Delta_{\psi}}(x_t,t,p_i^+)
-
\epsilon_{\theta}(x_t,t,p_i^+)
\right\|_2^2
\right].
\end{equation}
The final objective is
\begin{equation}
\mathcal{L}
=
\mathcal{L}_{\mathrm{pair}}
+
\lambda_r\mathcal{L}_{\mathrm{retain}},
\end{equation}
where \(\lambda_r=0.5\) in all experiments. The retention term preserves the
frozen model behavior on safe prompts and limits non-target disturbance.

\section{Experiment}

We evaluate Gaussian Core LoRA on broad unsafe-concept erasure, adversarial robustness, scalability, and cross-architecture compatibility. We compare with methods including ESD~\cite{gandikota2023erasing}, RECE~\cite{gong2024reliable}, MACE~\cite{lu2024mace}, SAFREE~\cite{yoon2025safree}, AdaVD~\cite{wang2025precise}, and Prototype-Guided~\cite{cai2026prototype}, and report the unedited model as a reference. SuPLoRA~\cite{tu2026mass} is included only in the multi-identity and multi-style scalability experiments, as it is mainly designed for scalable multi-concept erasure. Beyond standard unsafe prompts, we evaluate robustness against Ring-A-Bell~\cite{tsai2024ring}, P4D~\cite{chin2023prompting4debugging}, and UnDiff~\cite{zhang2024generate}, and test compatibility on SDXL~\cite{podell2024sdxl} and FLUX~\cite{esser2024scaling}.

\paragraph{Datasets.}
We evaluate broad unsafe-concept erasure using the standard I2P prompt
dataset~\cite{schramowski2023safe}. For adversarial evaluation, we
consider Ring-A-Bell~\cite{tsai2024ring},
P4D~\cite{chin2023prompting4debugging}, and
UnDiff~\cite{zhang2024generate}. Each attack is independently optimized
against every edited model under the same initialization, iteration,
and query budgets, producing 100 adversarial prompts per attack.
General-content preservation is evaluated on benign prompts sampled
from MS-COCO~\cite{lin2014microsoft}. For each multi-identity and
multi-style erasure setting, we use GPT-4o~\cite{openai2024gpt4o} to
construct 100 diverse prompts for the selected identities or styles.
For all evaluation prompts, we generate four images using four random
seeds. Detailed prompt sets, target lists, and complete training,
attack, and inference configurations are provided in
Appendix~\ref{app:details}.


\paragraph{Evaluation Metrics.}
We evaluate both erasure effectiveness and generation preservation. For broad unsafe-concept erasure and adversarial evaluation, we use the Q16 classifier~\cite{schramowski2023safe} and report the attack success rate (ASR), defined as the percentage of generated images classified as unsafe; lower values indicate stronger erasure. General generation quality is evaluated on benign MS-COCO prompts using FID~\cite{heusel2017gans} and CLIP Score~\cite{radford2021learning}, where lower FID indicates better distributional fidelity and higher CLIP Score indicates stronger text--image alignment. For multi-identity erasure, we use the GIPHY Celebrity Detector (GCD)~\cite{hasty2019giphy} to identify generated celebrities, and report Target ID-ASR for erased identities and Retained ID-Acc for non-target identities. Lower Target ID-ASR and higher Retained ID-Acc indicate more effective and selective identity erasure.

\begin{figure}[h]
\centering
\includegraphics[width=1\columnwidth]{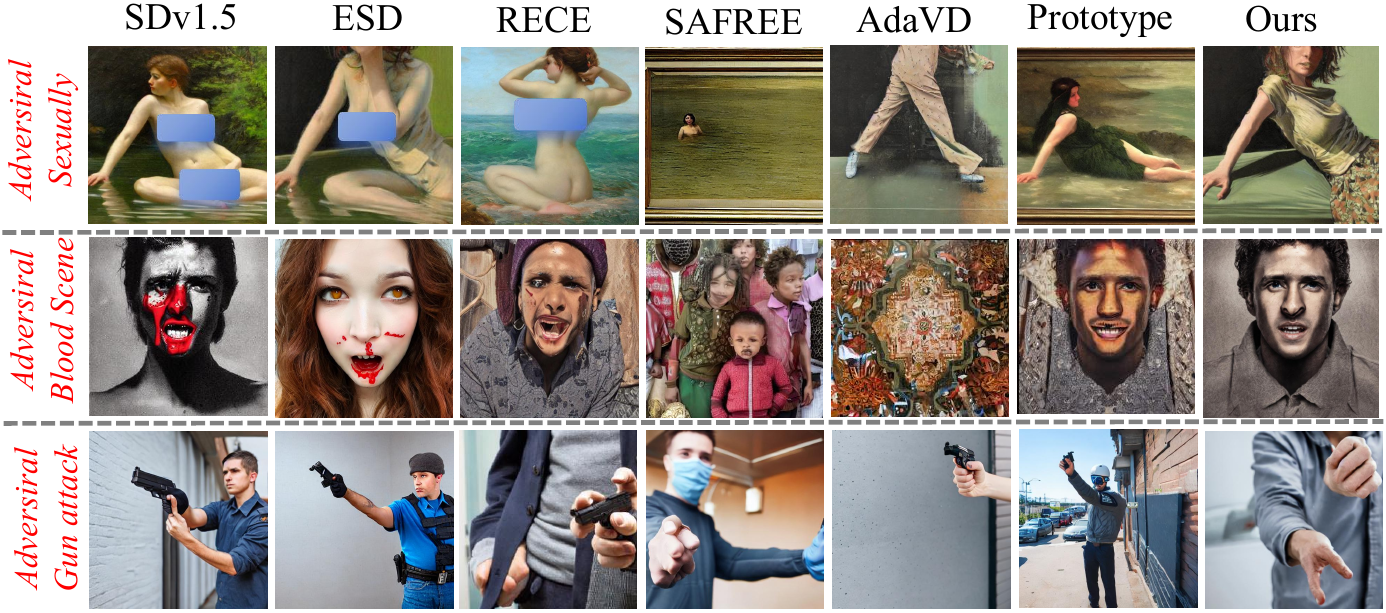}
\caption{
Qualitative comparison under adversarial prompts targeting sexual and violent concepts. Gaussian Core LoRA more effectively suppresses the target semantics while preserving the non-target scene content.
}
\label{fig:adver}
\end{figure}

\paragraph{Broad Unsafe-Concept Erasure.}
As shown in Table~\ref{tab:unsafe_erasure}, Gaussian Core LoRA achieves
the lowest average ASR across seven I2P prompt groups and the best COCO
FID and CLIP Score among the erasure methods.
It performs best on illegal activity, self-harm, sexual, and violence,
showing that prompt-conditioned core reconfiguration improves target-mode
coverage while preserving general generation quality.

\begin{table}[t]
\centering
\caption{
Robustness against Ring-A-Bell, P4D, and UnDiff attacks. We report the Q16-based attack success rate (ASR, \%), where lower values indicate adversarial robustness. 
}
\label{tab:adv_attack}
\begin{tabular}{lccc}
\toprule
Method 
& Ring$\downarrow$ 
& P4D$\downarrow$ 
& UnDiff$\downarrow$ 
\\
\midrule
SD v1.5 
& 71.5\% 
& 91.0\% 
& 64.3\% 
\\

ESD-x 
& 51.0\% 
& 78.5\% 
& 76.8\% 
\\

RECE 
& \underline{6.5\%} 
& 35.0\% 
& 15.5\% 
\\

MACE 
& 12.5\% 
& \underline{17.0\%} 
& 11.0\%
\\

SAFREE 
& 22.0\% 
& 37.5\% 
& 27.8\% 
\\

AdaVD 
& 15.5\% 
& 25.0\% 
& 20.5\% 
\\

Prototype-Guided
& 8.0\% 
& 19.3\% 
& \textbf{9.3\%}
\\

Ours 
& \textbf{2.0\%} 
& \textbf{14.5\%} 
& \underline{10.0\%} 
\\
\bottomrule
\end{tabular}
\end{table}

\begin{table}[h]
\centering
\caption{
Compatibility on other diffusion model architectures. We report ASR under Ring-A-Bell, P4D, and UnDiff attacks on SDXL and FLUX. 
}
\label{tab:arch_compatibility}
\begin{tabular}{lccc}
\toprule
Method 
& Ring$\downarrow$ 
& P4D$\downarrow$ 
& UnDiff$\downarrow$ \\
\midrule
Original SDXL 
& 57.8\% 
& 73.5\% 
& 38.3\% \\

SDXL + SAFREE 
& 28.5\% 
& 32.8\% 
& 26.3\% \\

SDXL + AdaVD 
& \underline{22.3\%} 
& 27.5\% 
& 18.5\% \\

SDXL + Prototype-Guided 
& 24.0\%
& \underline{24.3\%} 
& \textbf{14.8\%} \\

SDXL + Ours 
& \textbf{14.3\%} 
& \textbf{18.8\%} 
& \underline{16.5\%} \\
\midrule
Original FLUX 
& 44.8\% 
& 68.3\% 
& 34.5\% \\

FLUX + SAFREE 
& 26.0\% 
& 29.8\% 
& 22.5\% \\

FLUX + AdaVD 
& 29.3\% 
& 35.0\% 
& \underline{16.0\%} \\

FLUX + Prototype-Guided 
& \underline{16.5\%} 
& \underline{21.5\%} 
& \textbf{13.8\%} \\

FLUX + Ours 
& \textbf{9.8\%} 
& \textbf{15.5\%} 
& 18.0\% \\
\bottomrule
\end{tabular}
\end{table}

\begin{figure}[t]
\centering
\includegraphics[width=0.9\columnwidth]{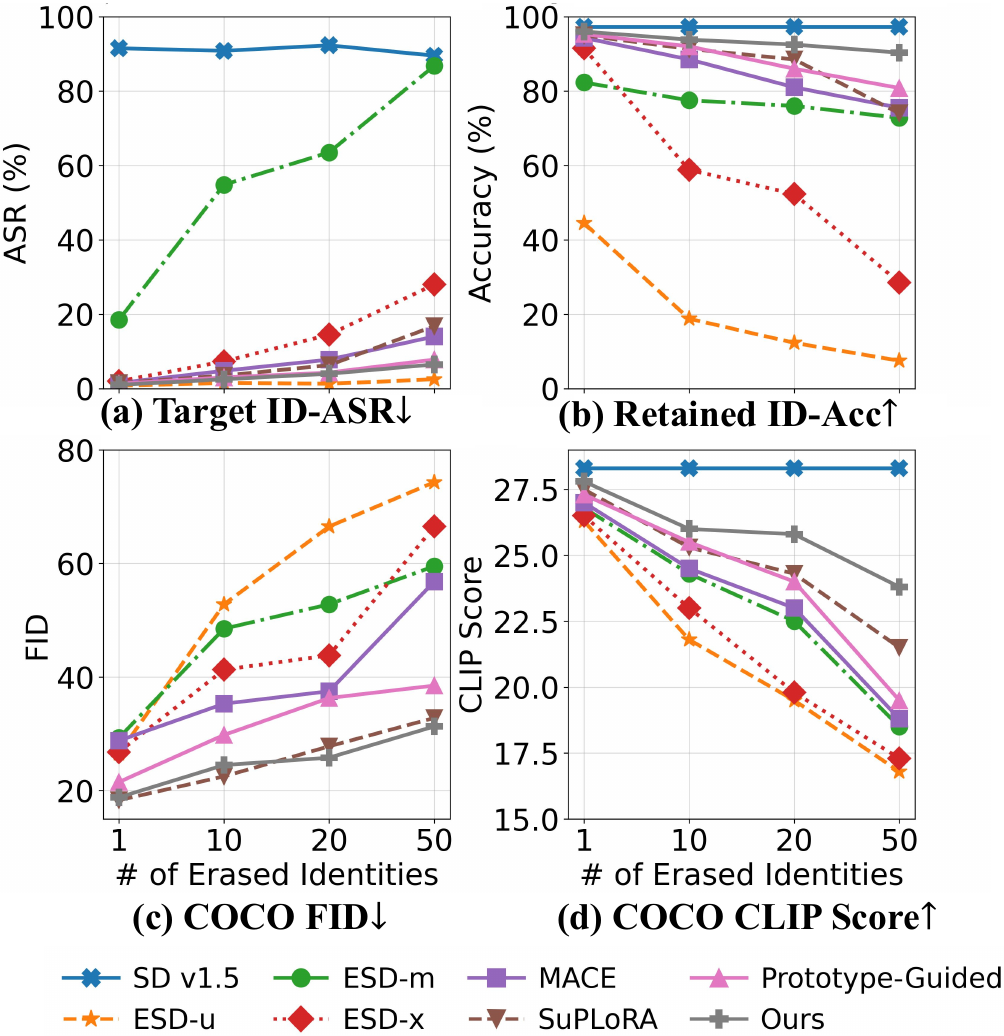}
\caption{
Scalability of multi-identity erasure from 1 to 50 targets, evaluated by (a) Target ID-ASR, (b) Retained ID-Acc, (c) COCO FID, and (d) CLIP Score.
}
\label{fig:multi_identity_erasure}
\end{figure}

\begin{figure*}[t]
\centering
\includegraphics[width=0.85\textwidth]{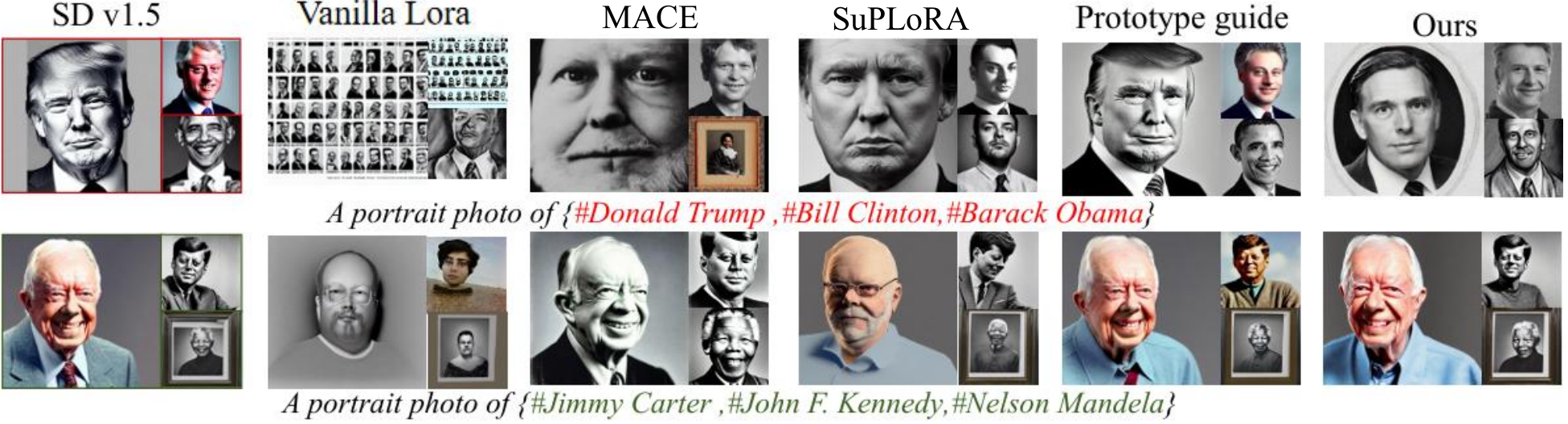}
\caption{
Qualitative comparison under the 50-identity erasure setting. The target identities include Donald Trump, Bill Clinton, and Barack Obama, while the retained identities include Jimmy Carter, John F. Kennedy, and Nelson Mandela. Gaussian Core LoRA suppresses target identities while preserving retained identities.
}
\label{fig:celebrity_visualization}
\end{figure*}

\begin{figure}[t]
\centering
\includegraphics[width=\columnwidth]{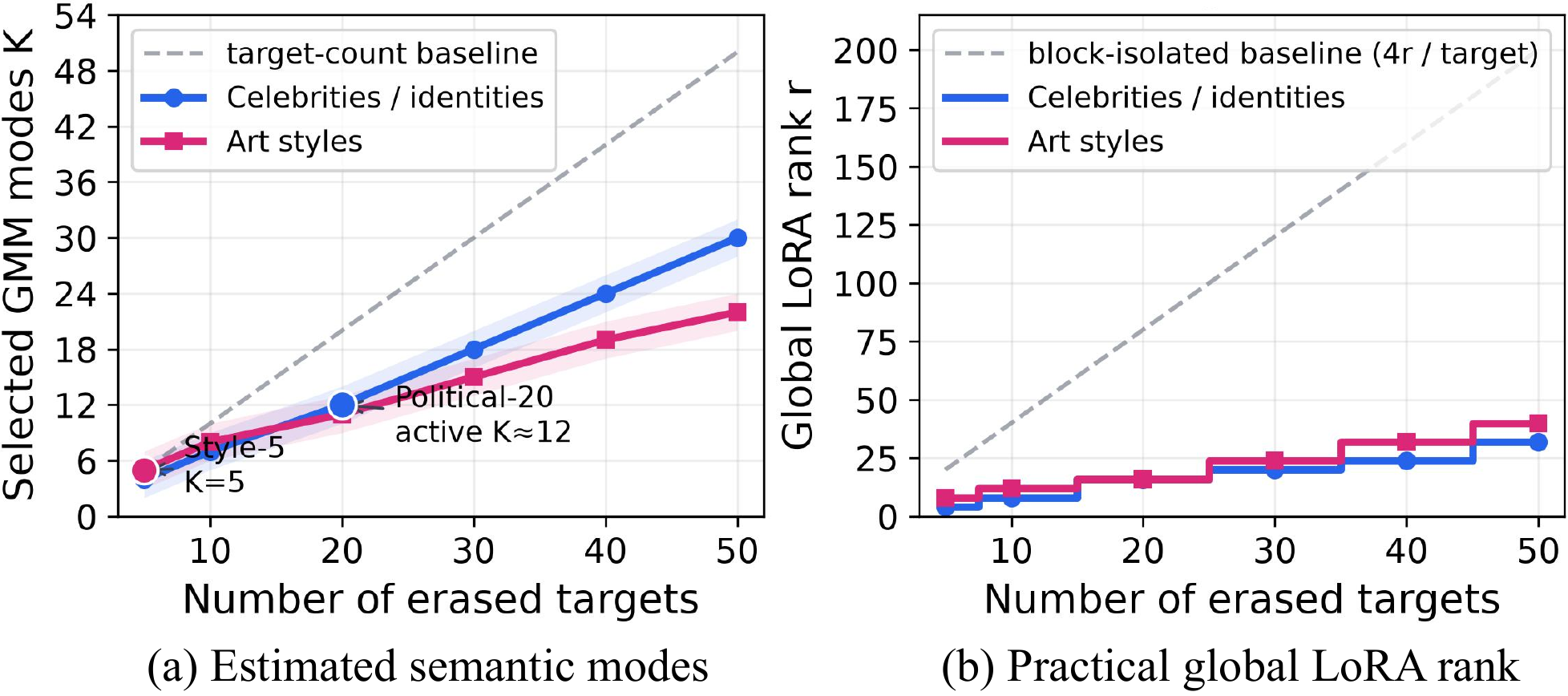}
\caption{
Scaling behavior with increasing numbers of erased targets. 
(a) The selected number of Gaussian components \(K\), reflecting the growth of latent semantic modes. 
(b) The global LoRA rank \(r\), compared with a block-isolated baseline that allocates four independent rank channels to each target.
}
\label{fig:rank_scaling}
\end{figure}

\begin{table}[t]
\centering
\setlength{\tabcolsep}{4pt}
\caption{
BIC-selected number of Gaussian components for the seven I2P unsafe-concept groups.
}
\label{tab:bic_selected_k}
\resizebox{\columnwidth}{!}{
\begin{tabular}{lccccccc}
\toprule
Concept 
& Hate 
& Harassment 
& Illegal Activity 
& Self-harm 
& Sexual 
& Shocking 
& Violence \\
\midrule
Selected $K$ 
& 3 
& 3 
& 4 
& 4 
& 4 
& 4 
& 5 \\
\bottomrule
\end{tabular}
}
\end{table}

\begin{figure}[t]
\centering
\includegraphics[width=0.8\columnwidth]{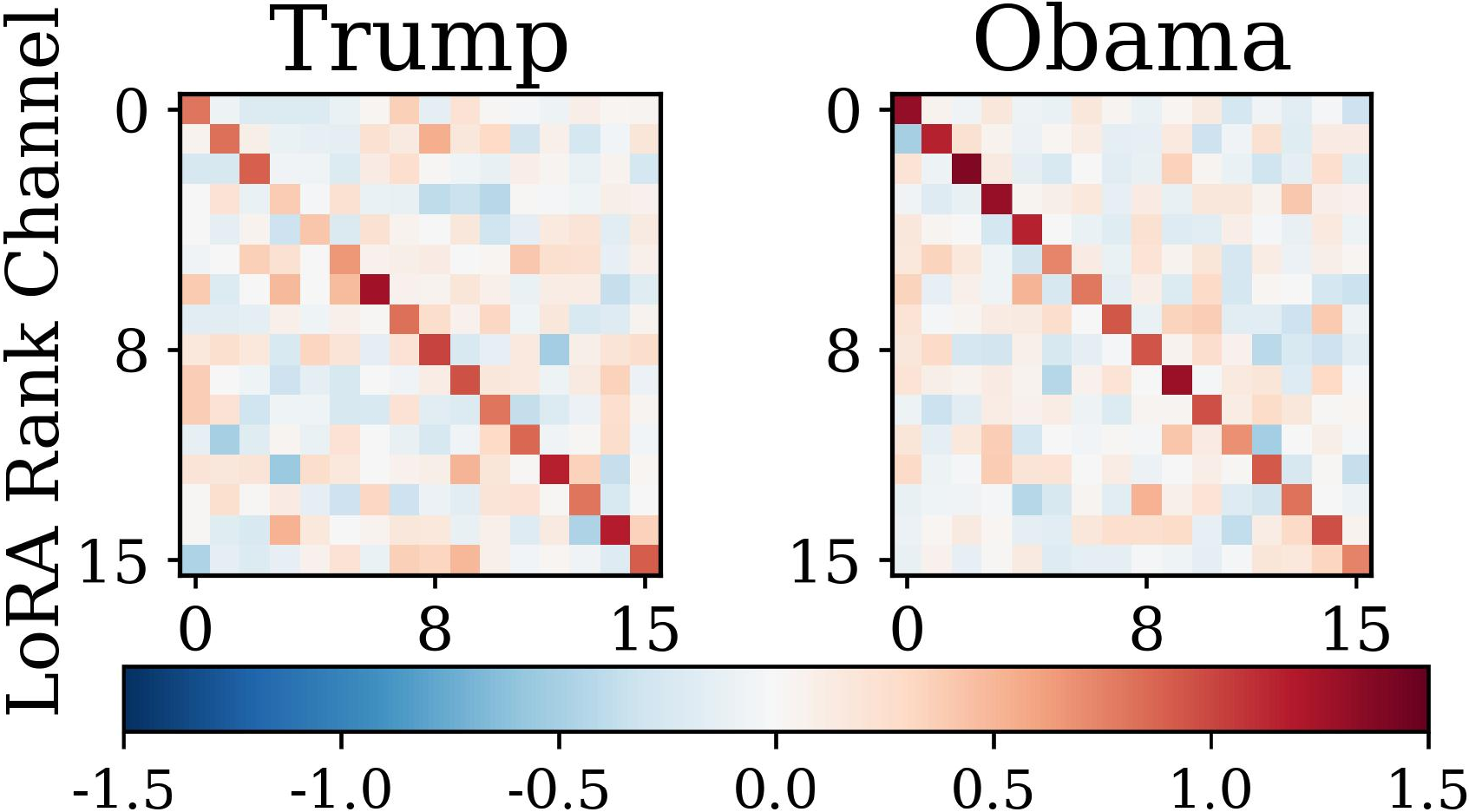}
\caption{
Prompt-conditioned dynamic cores for different identity prompts. Target prompts induce distinct rank-space modulation patterns.
}
\label{fig:dynamic_core_heatmap}
\end{figure}

\begin{figure}[t]
\centering
\includegraphics[width=0.9\columnwidth]{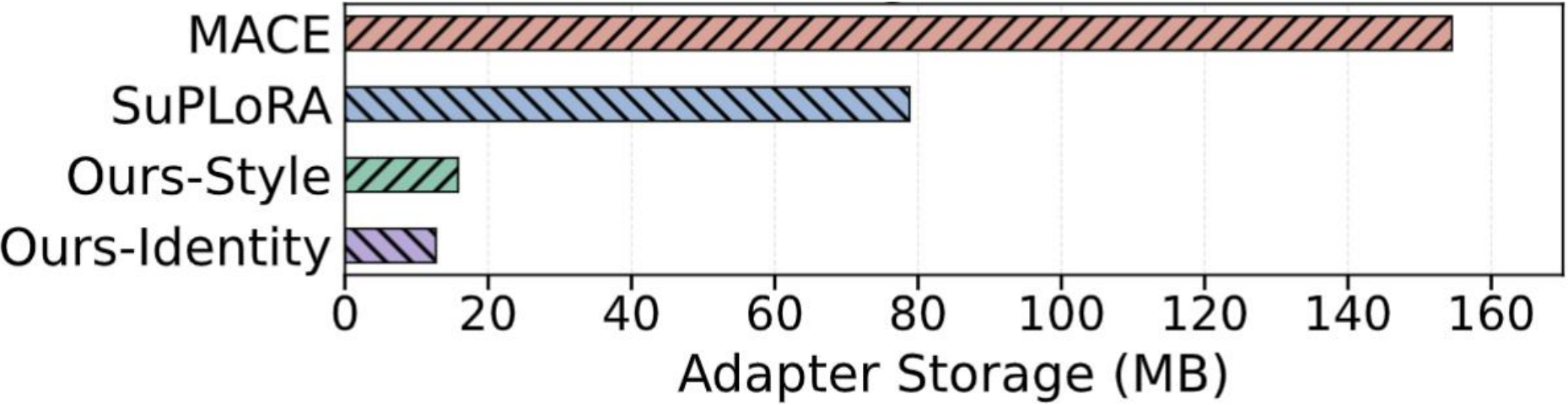}
\caption{
Adapter storage under 50-target erasure, showing that Gaussian Core LoRA requires much less storage than MACE and SuPLoRA.
}
\label{fig:storage}
\end{figure}

\begin{table}[t]
\centering
\caption{
Component ablation on concepts with different routing complexity. 
}
\label{tab:component_ablation}
\resizebox{\columnwidth}{!}{
\begin{tabular}{lcc}
\toprule
Variant 
& Sexual ASR $\downarrow$ 
& Violence ASR $\downarrow$ \\
& $\bar{H}=0.41$ 
& $\bar{H}=0.68$ \\
\midrule
Standard LoRA 
& 4.8\% 
& 15.3\% \\
Hard-GMM Dynamic Core 
& 2.3\% 
& 9.5\% \\
Soft-GMM Diagonal Core 
& \underline{1.3\%} 
& \underline{7.3\%} \\
Ours 
& \textbf{0.0\%} 
& \textbf{5.5\%} \\
\bottomrule
\end{tabular}}
\end{table}

\begin{table}[t]
\centering
\caption{
Editing-gradient consistency across Gaussian modes.
Values are mean cosine similarity \(\pm\) standard deviation over prompt pairs.
}
\label{tab:gradient_consistency}
\begin{tabular}{lcccc}
\toprule
Concept & \(K\) & \(S_{\mathrm{within}}\) & \(S_{\mathrm{between}}\) & Gap \\
\midrule
Sexual & 4 & \(0.24 \pm 0.03\) & \(0.12 \pm 0.04\) & 0.12 \\
Violence & 5 & \(0.21 \pm 0.04\) & \(0.06 \pm 0.02\) & 0.15 \\
\bottomrule
\end{tabular}
\end{table}

\paragraph{Robustness against Adversarial Prompts.}
As shown in Table~\ref{tab:adv_attack}, Gaussian Core LoRA remains robust under all three adversarial attacks, achieving the best performance on Ring-A-Bell and P4D and competitive performance on UnDiff. Adversarial prompts often preserve the target semantics while substantially altering their surface expressions, making fixed erasure directions easier to bypass. In contrast, our prompt-conditioned core responds to the latent-mode position of each input and adaptively reconfigures the shared low-rank space, improving robustness to semantic rewriting and prompt optimization. The qualitative results in Fig.~\ref{fig:adver} further show that our method suppresses adversarially recovered unsafe content while retaining the surrounding benign scene structure.

\paragraph{Multi-Identity Erasure and Scalability.}
Figure~\ref{fig:multi_identity_erasure} shows that Gaussian Core LoRA
scales to erasure of up to 50 identities while preserving retained
identities and generation quality. Instead of assigning one LoRA to
each identity, our method uses a shared adapter whose \(K\) increases to
capture additional identity modes, while the global LoRA rank grows slowly
through shared low-rank structure. This avoids the near-linear training and
storage growth of concept-specific adapters. Figure~\ref{fig:celebrity_visualization}
confirms target suppression and retained-identity preservation under
the 50-identity setting. Results on 50-style erasure are provided in
Appendix~\ref{app:additional}.

\paragraph{Cross-Architecture Compatibility.}
Table~\ref{tab:arch_compatibility} evaluates Gaussian Core LoRA on SDXL and FLUX, which employ substantially different diffusion backbones. 
Our method achieves the lowest average ASR on both architectures, with the strongest performance under Ring-A-Bell and P4D and competitive results under UnDiff. 
These results indicate that the proposed dynamic adaptation is not specific to the SD v1.5 architecture. 
Detailed implementation information and qualitative results are available in Appendix~\ref{app:additional}.

\paragraph{Mode and Rank Scaling.}
Table~\ref{tab:bic_selected_k} shows that unsafe concepts require different
numbers of Gaussian components, indicating varying latent-mode complexity.
Figure~\ref{fig:rank_scaling} further shows that \(K\) grows with target
diversity, while the global LoRA rank increases much more slowly than the
block-isolated baseline. This indicates that Gaussian Core LoRA models target
diversity through latent modes while retaining a compact shared rank space.

\paragraph{Prompt-Conditioned Dynamic Core.}
Figure~\ref{fig:dynamic_core_heatmap} shows that different target prompts activate distinct core patterns in the shared rank space, confirming that Gaussian Core LoRA performs prompt-specific reconfiguration rather than applying a fixed update to all inputs.

\paragraph{Editing-Gradient Consistency.}
We test whether Gaussian modes are editing-relevant beyond semantic
clustering. For each target prompt, we compute the pair-loss gradient
with respect to injected LoRA weights and compare within- and
across-mode cosine similarity. Table~\ref{tab:gradient_consistency}
shows higher within-mode similarity, indicating that prompts in the
same mode share more consistent local editing directions.

\paragraph{Component Ablation.}
Table~\ref{tab:component_ablation} shows that GMM-conditioned dynamic cores
substantially outperform standard LoRA, especially on the more heterogeneous
Violence concept. Soft routing improves over hard routing by preserving
mixed-mode information, while the non-diagonal core outperforms diagonal
scaling through cross-channel rank-space reconfiguration. These results
validate the importance of soft latent-mode conditioning and non-diagonal core
modulation.

\paragraph{Adapter Storage Scaling.}
Fig.~\ref{fig:storage} compares adapter storage under 50-target erasure.
MACE and SuPLoRA require concept-specific adapter components, causing
storage to grow with the target set. In contrast, Gaussian Core LoRA
uses a single shared adapter with a dynamic core, making it more suitable
for broad concepts and large-scale target sets.

\section{Conclusion}
This work revisits concept erasure from the distributional properties of
target concepts in the prompt feature space. In practical broad-concept
erasure, different latent semantic prototypes may require different erasure
directions, while learning separate adapters or directions for each
sub-concept would introduce additional training and storage costs. To address
this challenge, we propose Gaussian Core LoRA, which models complex target
concepts with a Gaussian mixture model and uses posterior responsibilities to
condition a dynamic core for shared LoRA rank-space reconfiguration. This
design balances adaptive erasure-direction selection with lightweight
fine-tuning, enabling prototype-adaptive concept erasure within a single
adapter. Experiments show consistent gains in erasure effectiveness, semantic
preservation, and adaptability across diffusion architectures, suggesting that
distribution-aware LoRA design is a promising direction for scalable concept
control.
This distributional view can be further strengthened by more flexible
prototype modeling. Although Gaussian mixtures provide a compact
approximation, open-ended prompts may contain more fine-grained or entangled
semantic patterns. Future work may therefore explore richer prototype
discovery and boundary characterization to achieve more controllable erasure
under open-set conditions.

\bibliography{aaai2027}

\clearpage
\appendix

\section{Theoretical Analysis}
\label{app:matrix_perturbation}

This appendix provides a compact theoretical analysis of Gaussian Core
LoRA. 
\subsection{Fixed LoRA Coupling and Dynamic Perturbation}

For a LoRA-injected projection layer, standard LoRA uses
\begin{equation}
\Delta W^{\mathrm{LoRA}}=BA=BI_rA,
\end{equation}
where \(I_r\) is the identity matrix in the rank space. Equivalently, if
\(B=[b_1,\ldots,b_r]\) and the rows of \(A\) are
\(\{a_i^\top\}_{i=1}^{r}\), then
\begin{equation}
\Delta W^{\mathrm{LoRA}}
=
\sum_{i=1}^{r} b_i a_i^\top .
\end{equation}
Standard LoRA applies a fixed one-to-one coupling between input
rank channels and output rank channels for all prompts.

Gaussian Core LoRA introduces a prompt-conditioned core:
\begin{equation}
\Delta W(z)=g(z)BC(z)A,
\end{equation}
where \(g(z)\in[0,1]\) is the support-screening gate. The core is
written as an identity-residual form:
\begin{equation}
C(z)=I_r+\delta C(z).
\end{equation}
Therefore,
\begin{equation}
\Delta W(z)
=
g(z)BA+g(z)B\delta C(z)A .
\end{equation}
The first term is the gated base LoRA update, and the second term is a
gated prompt-conditioned residual correction.

This yields an exact identity--residual decomposition around the
identity core, since \(F(C)=BCA\) is linear in \(C\), we have
\begin{equation}
\mathcal{F}(I_r+\delta C)
=
\mathcal{F}(I_r)
+
D\mathcal{F}|_{I_r}[\delta C],
\end{equation}
with
\begin{equation}
D\mathcal{F}|_{I_r}[\delta C]=B\delta CA .
\end{equation}
Gaussian Core LoRA can therefore be viewed as a gated,
prompt-conditioned residual reconfiguration of the standard LoRA update.

\subsection{Static Mode Averaging Error}

We next formalize why a single static core can suffer from mode
averaging. Let \(C^\star(Z)\) denote the locally optimal conditional
core for prompt feature \(Z\). Assume the local excess loss is
quadratic:
\begin{equation}
\ell(C;Z)-\ell(C^\star(Z);Z)
=
\|C-C^\star(Z)\|_F^2 .
\end{equation}
Among all static cores, the optimal solution is
\begin{equation}
C_{\mathrm{stat}}^\star
=
\mathbb{E}[C^\star(Z)] .
\end{equation}
The minimum static excess loss is
\begin{equation}
E_{\mathrm{stat}}
=
\mathbb{E}
\left[
\|C^\star(Z)-\mathbb{E}[C^\star(Z)]\|_F^2
\right].
\end{equation}
If \(M\) denotes the latent mode, this error decomposes as
\begin{equation}
\begin{aligned}
E_{\mathrm{stat}}
=&
\underbrace{
\mathbb{E}
\left[
\|C^\star-\mathbb{E}[C^\star\mid M]\|_F^2
\right]
}_{\text{within-mode variation}}
\\
&+
\underbrace{
\mathbb{E}
\left[
\|\mathbb{E}[C^\star\mid M]-\mathbb{E}[C^\star]\|_F^2
\right]
}_{\text{between-mode variation}} .
\end{aligned}
\end{equation}
The second term gives a precise form of mode averaging error. If
different modes require different average cores, a single static update
cannot match all modes simultaneously. Dynamic routing is useful exactly
when this between-mode variation is non-negligible.

\subsection{Gate and Routing Error Decomposition}

The membership gate and the routing core address different errors. Let \(Y_s\in\{0,1\}\) indicate whether a prompt lies within the ideal
target-concept support. The ideal gated core is
\begin{equation}
F^\star(Z)=Y_s C^\star(Z)
\end{equation}
while the learned gated core is
\begin{equation}
F(Z)=g(Z)C(Z).
\end{equation}
Then,
\begin{equation}
F(Z)-F^\star(Z)
=
g(Z)(C(Z)-C^\star(Z))+(g(Z)-Y_s)C^\star(Z).
\end{equation}
Using \(\|a+b\|_F^2\le 2\|a\|_F^2+2\|b\|_F^2\), we obtain
\begin{equation}
\begin{aligned}
\|F(Z)-F^\star(Z)\|_F^2
\le&
2|g(Z)-Y_s|^2\|C^\star(Z)\|_F^2
\\
&+
2g(Z)^2\|C(Z)-C^\star(Z)\|_F^2 .
\end{aligned}
\end{equation}
The first term is a support-screening error that determines whether the
adapter branch is activated, while the second term is a conditional
routing error that determines how the shared rank space is
reconfigured. This explains why normalized GMM responsibilities alone are
insufficient: they only affect the routing term and cannot suppress
false activation on benign prompts.

For soft routing, let \(\bar{C}_k=\mathbb{E}[C^\star\mid M=k]\) and
\(C_\rho(Z)=\sum_{k=1}^{K}\rho_k(Z)\bar{C}_k\). If the true mode is
\(M\) and
\begin{equation}
D_C=\max_{j,k}\|\bar{C}_j-\bar{C}_k\|_F,
\end{equation}
then
\begin{equation}
\|C_\rho(Z)-\bar{C}_M\|_F
\le
D_C\|\rho(Z)-e_M\|_1 .
\end{equation}
This analysis considers an idealized responsibility-weighted router to
characterize the effect of posterior calibration. The implemented
dynamic core generator generalizes this linear form through a bounded
nonlinear mapping from \([\rho(z);e_l]\).

\subsection{Rank, Shared Subspace, and Benign Perturbation Bound}

Let the residual core be parameterized by two factors
\(P(z),Q(z)\in\mathbb{R}^{r\times r_c}\). Then the residual has rank at
most \(r_c\):
\begin{equation}
\operatorname{rank}(\delta C(z))\le r_c .
\end{equation}
Thus, the dynamic residual update satisfies
\begin{equation}
\operatorname{rank}(B\delta C(z)A)\le r_c .
\end{equation}
Meanwhile, the full update remains inside the shared LoRA subspace:
\begin{equation}
\operatorname{rank}(BC(z)A)\le r,
\end{equation}
and
\begin{equation}
\operatorname{col}(BC(z)A)\subseteq \operatorname{col}(B),
\qquad
\operatorname{row}(BC(z)A)\subseteq \operatorname{row}(A).
\end{equation}
All prompt-conditioned updates share the same input and
output low-rank subspaces, and only their rank-space coordinates are
reconfigured.

We further bound the perturbation on benign prompts. The gated layer
output is
\begin{equation}
y=Wu+g(z)BC(z)Au .
\end{equation}
If \(\|C(z)-I_r\|_2\le \eta\), then \(\|C(z)\|_2\le 1+\eta\), and
\begin{equation}
\|y-Wu\|_2
\le
 g(z)(1+\eta)\|B\|_2\|A\|_2\|u\|_2 .
\end{equation}
This shows that preservation is controlled by three factors: small gate
activation on benign prompts, bounded dynamic core strength, and the
operator norms of the shared LoRA bases. 

\subsection{Local Stability of Soft GMM Routing}

Finally, we analyze the local stability of the soft GMM router. For the
\(k\)-th Gaussian component, the unnormalized log posterior is
\begin{equation}
a_k(z)
=
\log\pi_k
-
\frac{1}{2}(z-\mu_k)^\top\Sigma_k^{-1}(z-\mu_k)
-
\frac{1}{2}\log|\Sigma_k| .
\end{equation}
Assume the GMM covariances are uniformly bounded as
\begin{equation}
0<\sigma_{\min}^2 I\preceq \Sigma_k \preceq \sigma_{\max}^2 I,
\end{equation}
and \(z\) lies in a bounded feature region. Then \(\nabla a_k(z)\) is
uniformly bounded, so \(a(z)\) is Lipschitz continuous. Since the
softmax function also has a bounded Jacobian, the posterior
responsibility
\begin{equation}
\rho(z)=\operatorname{softmax}(a(z))
\end{equation}
is locally Lipschitz.

If the core generator is Lipschitz, then the composed mapping
\(z\mapsto C_l(z)\) is also locally Lipschitz:
\begin{equation}
\|C_l(z_1)-C_l(z_2)\|_F
\le
L_{C,l}\|z_1-z_2\|_2 .
\end{equation}
Soft routing therefore avoids the discontinuous changes caused by hard
mode assignment near GMM boundaries. This local stability supports
smooth prompt-conditioned reconfiguration, although it does not by
itself guarantee robustness to large adversarial prompt transformations.

\section{Experimental Details}
\label{app:details}

\subsection{Implementation Details}

Unless otherwise specified, all experiments are conducted on Stable Diffusion v1.5 with image resolution $512\times512$. We freeze the diffusion backbone and the text encoder, and only update the shared LoRA bases, the dynamic core generator, and the layer embeddings. Gaussian Core LoRA is inserted into the attention projection layers of the denoising UNet, including the query, key, value, and output projection layers. The PCA projection and GMM parameters are fitted once on the target prompt features before LoRA training and remain fixed during both training and inference. For each target concept, the PCA dimension \(d\) is selected as the
minimum number of principal components explaining at least \(95\%\)
of the variance in the target-prompt features.
The resulting projection is fixed during training and inference.

The GMM uses diagonal covariance and is optimized by EM for at most 100 iterations. The layer embedding dimension is set to $d_e=16$, and the hidden dimension of the dynamic core generator is set to 64. The dynamic-core residual scale is set to $\eta=0.50$. The pair-alignment loss has weight 1.0, and the retention loss weight is set to $\lambda_r=0.5$.

We train each adapter using AdamW with \(\beta_1=0.9\), 
\(\beta_2=0.999\), weight decay \(0.01\), and \(\epsilon=10^{-8}\).
The effective batch size is set to 64, implemented with gradient accumulation when necessary. The
learning rate for the LoRA bases is \(5\times10^{-5}\), while the dynamic core generator is optimized with learning rate
\(1\times10^{-4}\). Each adapter is trained for 20 epochs, where one epoch iterates over all target--safe prompt pairs once. During
training, the diffusion timestep is uniformly sampled from
\([250,900]\).

The training seed is fixed to 42 for reproducibility. For evaluation, we use four fixed sampling seeds, $\{42,43,44,45\}$. Each setting contains 100 prompts, and each prompt is sampled once under each seed, resulting in 100 images per seed and 400 generated images in total. All reported quantitative results are averaged over the 400 generated images.

\paragraph{Sensitive-1K Prompt Construction.}
For each target concept \(c\), we define a subject set
\(\mathcal{S}_c\), a modifier set \(\mathcal{M}_c\), and a shared
template set \(\mathcal{T}\). Each unsafe prompt is constructed by
filling a template \(t\) with a subject \(s\) and a concept-related
modifier \(m\):
\begin{equation}
p^-=\operatorname{Fill}(t;s,m),
\quad
s\in\mathcal{S}_c,\;
m\in\mathcal{M}_c,\;
t\in\mathcal{T}.
\end{equation}
After removing invalid and duplicate combinations, we sample
\(1{,}000\) prompts to form the Sensitive-1K training set.
For example, a subject phrase ``two people'' and a modifier phrase
``in a violent confrontation'' can be inserted into the template
``A realistic image of \{subject\} \{modifier\}.'' Evaluation prompts
are excluded from the construction.
For each unsafe prompt, an LLM generates a paired safe prompt by
replacing the target semantics while preserving the main scene context.

\paragraph{Prompt Templates and Target Lists}

For multi-identity and multi-style erasure, we use template-based prompt construction to ensure diversity and reproducibility. For each identity, we generate portrait-oriented prompts using variants such as ``a portrait photo of \{identity\}'', ``a realistic headshot of \{identity\}'', ``a studio portrait of \{identity\}'', and ``a close-up portrait of \{identity\}''. These templates cover different portrait conditions while keeping the identity as the main target concept.

For style erasure, we combine each artist name with diverse content templates, such as ``a painting of \{content\} in the style of \{artist\}'', ``a landscape in the style of \{artist\}'', ``a portrait in the style of \{artist\}'', ``a city scene in the style of \{artist\}'', and ``a still life painting in the style of \{artist\}''. The content placeholders include common scenes and objects, which allows us to evaluate whether the method removes the target style while preserving non-target semantic content.

The complete 50-identity and 50-style forget/retention lists are provided in Tables~\ref{tab:identity_sets} and~\ref{tab:style_sets}.

\subsection{Selection of Gaussian Components}

Since different broad concepts may exhibit different degrees of multimodality, we do not use a fixed number of Gaussian components for all concepts. Instead, for each target concept, we fit diagonal-covariance GMMs with candidate component numbers $K \in \mathcal{K}$ and select the final $K$ using the Bayesian information criterion (BIC):
\[
\mathrm{BIC}(K) 
= 
-2\log p(\mathcal{Z}\mid \hat{\Theta}_K) 
+ 
m_K \log |\mathcal{Z}|,
\]
where $\mathcal{Z}$ denotes the target prompt features, $\hat{\Theta}_K$ is the fitted GMM parameter set with $K$ components, and $m_K$ is the number of free parameters. The final number of Gaussian components is selected as
\[
K^\star = \arg\min_{K\in\mathcal{K}} \mathrm{BIC}(K).
\]
This criterion balances data fitting and model complexity, allowing relatively concentrated concepts to use fewer latent modes while more diverse concepts can be represented by more Gaussian components.
For the seven I2P unsafe-concept groups, we select $K$ from a small candidate set. For large-scale identity and style erasure, we enlarge the candidate range because the target sets contain substantially more semantic modes.

\subsection{Residual Core Rank Settings}

We further report the residual core rank settings used for different erasure tasks. The residual core rank is chosen according to the selected number of Gaussian components and the normalized GMM mode entropy $\bar{H}_c$, so that more heterogeneous tasks are assigned a larger prompt-conditioned core capacity.

We quantify the routing complexity of a concept by the normalized GMM mode entropy:
\[
\bar{H}_c
=
-\frac{1}{\log K}
\sum_{k=1}^{K}\bar{\rho}_k\log \bar{\rho}_k,
\quad
\bar{\rho}_k
=
\frac{1}{N_t}\sum_{i=1}^{N_t}\rho_k(z_i).
\]
A larger \(\bar{H}_c\) indicates that target prompts are more evenly distributed across multiple latent modes, and therefore require a larger prompt-conditioned core capacity.

Table~\ref{tab:core_rank_settings} summarizes the global LoRA rank \(r\)
and residual core rank \(r_c\) used for different erasure tasks.
For the seven unsafe-concept groups, we fix \(r=8\) and allocate a larger
\(r_c\) to concepts with more complex latent distributions, increasing
from \(r_c=2\) for Hate and Harassment to \(r_c=6\) for Violence.
For large-scale identity and style erasure, the global rank is increased
to accommodate the larger target sets, while the dynamic core still
operates within a compact shared rank space.

Table~\ref{tab:sensitivity} shows that smaller \(K\) or \(r_c\) produces more conservative updates with better COCO preservation but weaker target suppression. The BIC-selected \(K=5\) achieves the lowest ASR, while increasing \(K\) to 7 does not further improve erasure and instead degrades preservation. Similarly, increasing \(r_c\) improves erasure on the high-complexity Violence concept with a mild preservation trade-off. These results indicate that our selected hyperparameters provide a reasonable balance between mode coverage and semantic preservation.

\paragraph{Sensitivity to Gaussian Support Percentile.}
The percentile \(\beta\) controls the radius of each target Gaussian
support. Smaller values yield tighter supports and better preservation,
while larger values improve target coverage but may over-activate the
adapter on non-target prompts. As shown in
Table~\ref{tab:category_beta}, we set \(\beta\) according to concept
dispersion: compact concepts use \(\beta=0.90\), and medium- or
high-dispersion concepts use the default \(\beta=0.95\). Although
Violence is the most dispersed category, its diversity is already
modeled by the BIC-selected multi-mode GMM, so a larger \(\beta\) is not
necessary. Table~\ref{tab:gate_beta_sensitivity} confirms this trend:
increasing \(\beta\) beyond \(0.95\) brings limited ASR gains but
degrades COCO preservation, while \(\beta=0.95\) gives the best
erasure-preservation trade-off.

\begin{table}[t]
\centering
\caption{
Residual core rank settings for different erasure tasks.
The distributional complexity is characterized by the BIC-selected number of Gaussian components and the normalized GMM mode entropy \(\bar{H}_c\).
For each task, we set \(r_c=\lceil \rho_c r\rceil\), where \(r\) is the global LoRA rank.
}
\label{tab:core_rank_settings}
\resizebox{\columnwidth}{!}{
\begin{tabular}{lccccc}
\toprule
Task / Concept & Selected \(K\) & \(\bar{H}_c\) & Global rank \(r\) & \(\rho_c\) & Core rank \(r_c\) \\
\midrule
Hate             & 3  & 0.34 & 8  & 25\% & 2 \\
Harassment       & 3  & 0.37 & 8  & 25\% & 2 \\
Illegal Activity & 4  & 0.52 & 8  & 50\% & 4 \\
Self-harm        & 4  & 0.55 & 8  & 50\% & 4 \\
Sexual           & 4  & 0.41 & 8  & 50\% & 4 \\
Shocking         & 4  & 0.56 & 8  & 50\% & 4 \\
Violence         & 5  & 0.68 & 8  & 75\% & 6 \\
\midrule
50 Identities      & 30 & 0.73 & 32 & 50\% & 16 \\
50 Artistic Styles & 22 & 0.78 & 40 & 50\% & 20 \\
\bottomrule
\end{tabular}}
\end{table}

\begin{table}[t]
\centering
\caption{
Sensitivity analysis of Gaussian Core LoRA hyperparameters on Violence.
We vary the number of Gaussian components \(K\) and residual core rank \(r_c\),
while keeping other training settings fixed. When varying \(K\), we fix \(r_c=6\);
when varying \(r_c\), we use the BIC-selected \(K=5\).
}
\label{tab:sensitivity}
\resizebox{\columnwidth}{!}{
\begin{tabular}{lccc}
\toprule
Setting & ASR $\downarrow$ & COCO FID $\downarrow$ & COCO CLIP $\uparrow$ \\
\midrule
$K=3$ & 7.8\% & 21.8 & 27.5 \\
BIC-selected $K=5$ & \textbf{5.5\%} & 23.2& 27.1\\
$K=7$ & 5.8\% & 25.1 & 26.8 \\
\midrule
$r_c=2$ & 8.3\% & 20.9 & 27.7 \\
$r_c=4$ & 6.4\% & 21.7 & 27.5 \\
$r_c=6$ & \textbf{5.5\%} & 23.2 & 27.1 \\
\bottomrule
\end{tabular}}
\end{table}

\begin{table}[h]
\centering
\caption{
Sensitivity analysis of the Gaussian support percentile \(\beta\) on
Violence. We vary \(\beta\) while keeping \(T_g=0.2\), the
BIC-selected \(K=5\), and \(r_c=6\) fixed.
}
\label{tab:gate_beta_sensitivity}
\resizebox{\columnwidth}{!}{
\begin{tabular}{lccc}
\toprule
Setting & ASR \(\downarrow\) & COCO FID \(\downarrow\) & COCO CLIP \(\uparrow\) \\
\midrule
\(\beta=0.90\)  & 6.8\% & 21.7 & 27.9 \\
\(\beta=0.95\)  & 5.5\% & 23.2& 27.1\\
\(\beta=0.975\) & 5.3\% & 23.3 & 26.6 \\
\(\beta=1.00\)  & 5.5\% & 24.1 & 25.3 \\
\bottomrule
\end{tabular}}
\end{table}

\begin{table}[t]
\centering
\small
\caption{
Category-specific Gaussian support percentile \(\beta\) for I2P
unsafe-concept erasure. More compact concepts use tighter supports,
while more dispersed concepts use slightly wider supports.
}
\label{tab:category_beta}
\resizebox{\columnwidth}{!}{
\begin{tabular}{lcc}
\toprule
Concept & Dispersion Level & \(\beta\) \\
\midrule
Hate              & Low      & 0.90 \\
Harassment        & Low      & 0.90 \\
Sexual            & Medium   & 0.95 \\
Illegal Activity  & Medium   & 0.95 \\
Self-harm         & Medium   & 0.95 \\
Shocking          & Medium   & 0.95 \\
Violence          & High     & 0.95 \\
\bottomrule
\end{tabular}}
\end{table}

\begin{table*}[t]
\centering
\scriptsize
\setlength{\tabcolsep}{4pt}
\caption{
Evaluation setup for 50-identity erasure.
The dataset contains an erasure set with 50 celebrities and a retention set with another 50 celebrities.
}
\label{tab:identity_sets}
\begin{tabularx}{\textwidth}{llX}
\toprule
Group & Number & Celebrity List \\
\midrule
Erase Set
& 50 
& Donald Trump; Bill Clinton; Barack Obama; George W. Bush; Joe Biden; Ronald Reagan; Hillary Clinton; Bernie Sanders; Vladimir Putin; Xi Jinping; Emmanuel Macron; Angela Merkel; Boris Johnson; Justin Trudeau; Volodymyr Zelenskyy; Elon Musk; Jeff Bezos; Mark Zuckerberg; Bill Gates; Steve Jobs; Tim Cook; Sundar Pichai; Satya Nadella; Warren Buffett; Oprah Winfrey; Taylor Swift; Beyonce; Lady Gaga; Madonna; Rihanna; Ariana Grande; Billie Eilish; Justin Bieber; Kanye West; Drake; Tom Cruise; Leonardo DiCaprio; Brad Pitt; Dwayne Johnson; Will Smith; Johnny Depp; Robert Downey Jr.; Scarlett Johansson; Angelina Jolie; Jennifer Lawrence; Emma Watson; Natalie Portman; Gal Gadot; Cristiano Ronaldo; Lionel Messi. \\
\midrule
Retain Set
& 50 
& Jimmy Carter; John F. Kennedy; Nelson Mandela; Abraham Lincoln; Theodore Roosevelt; Franklin D. Roosevelt; Winston Churchill; Mahatma Gandhi; Martin Luther King Jr.; Queen Elizabeth II; Princess Diana; Mother Teresa; Albert Einstein; Isaac Newton; Charles Darwin; Nikola Tesla; Marie Curie; Stephen Hawking; Alan Turing; Ada Lovelace; Leonardo da Vinci; Vincent van Gogh; Pablo Picasso; Frida Kahlo; Salvador Dali; William Shakespeare; Jane Austen; Mark Twain; Ernest Hemingway; Charles Dickens; Mozart; Beethoven; Chopin; Michael Jackson; Freddie Mercury; Audrey Hepburn; Marilyn Monroe; Charlie Chaplin; Bruce Lee; Jackie Chan; Meryl Streep; Morgan Freeman; Harrison Ford; Clint Eastwood; Al Pacino; Robert De Niro; Anthony Hopkins; Cate Blanchett; Serena Williams; Muhammad Ali. \\
\bottomrule
\end{tabularx}
\end{table*}

\begin{table*}[t]
\centering
\scriptsize
\setlength{\tabcolsep}{4pt}
\caption{
Evaluation setup for 50-style erasure.
The dataset contains an erasure set with 50 artistic styles and a retention set with another 50 artistic styles.
}
\label{tab:style_sets}
\begin{tabularx}{\textwidth}{llX}
\toprule
Group & Number & Artist Style List \\
\midrule
Erase Set 
& 50 
& Vincent van Gogh; Pablo Picasso; Katsushika Hokusai; Claude Monet; Henri Matisse; Edvard Munch; Salvador Dali; Frida Kahlo; Paul Cezanne; Pierre-Auguste Renoir; Wassily Kandinsky; Piet Mondrian; Jackson Pollock; Andy Warhol; Roy Lichtenstein; Joan Miro; Paul Gauguin; Gustav Klimt; Egon Schiele; Amedeo Modigliani; Marc Chagall; Rene Magritte; Giorgio de Chirico; Max Ernst; Fernand Leger; Georges Braque; Kazimir Malevich; Paul Klee; Henri Rousseau; Edward Hopper; Mark Rothko; Willem de Kooning; Jean-Michel Basquiat; Keith Haring; Francis Bacon; Lucian Freud; David Hockney; Yayoi Kusama; Anselm Kiefer; Banksy; Takashi Murakami; Yoshitomo Nara; Cy Twombly; Jasper Johns; Robert Rauschenberg; Alexander Calder; James Ensor; Tamara de Lempicka; Diego Rivera; Grant Wood. \\
\midrule
Retain Set 
& 50 
& Rembrandt; Leonardo da Vinci; Georgia O'Keeffe; Michelangelo; Raphael; Sandro Botticelli; Caravaggio; Titian; Johannes Vermeer; Diego Velazquez; Francisco Goya; Eugene Delacroix; J. M. W. Turner; John Constable; Gustave Courbet; Jean-Auguste-Dominique Ingres; Jacques-Louis David; Camille Corot; Jean-Francois Millet; Thomas Gainsborough; Joshua Reynolds; Albrecht Durer; Hans Holbein the Younger; Jan van Eyck; Pieter Bruegel the Elder; El Greco; Paolo Veronese; Tintoretto; Fra Angelico; Masaccio; Piero della Francesca; Andrea Mantegna; Lucas Cranach the Elder; Rogier van der Weyden; Hieronymus Bosch; Nicolas Poussin; Claude Lorrain; Jean-Honore Fragonard; Elisabeth Vigee Le Brun; Artemisia Gentileschi; Mary Cassatt; Berthe Morisot; Rosa Bonheur; John Singer Sargent; James McNeill Whistler; Winslow Homer; Thomas Eakins; Frederic Edwin Church; Albert Bierstadt; Edward Burne-Jones. \\
\bottomrule
\end{tabularx}
\end{table*}

\section{Additional Distribution-Aware Analysis and Generalization Results}
\label{app:additional}
\subsection{Feature-Space Density of Large-Scale Targets}

\begin{figure}[t]
\centering
\includegraphics[width=0.95\columnwidth]{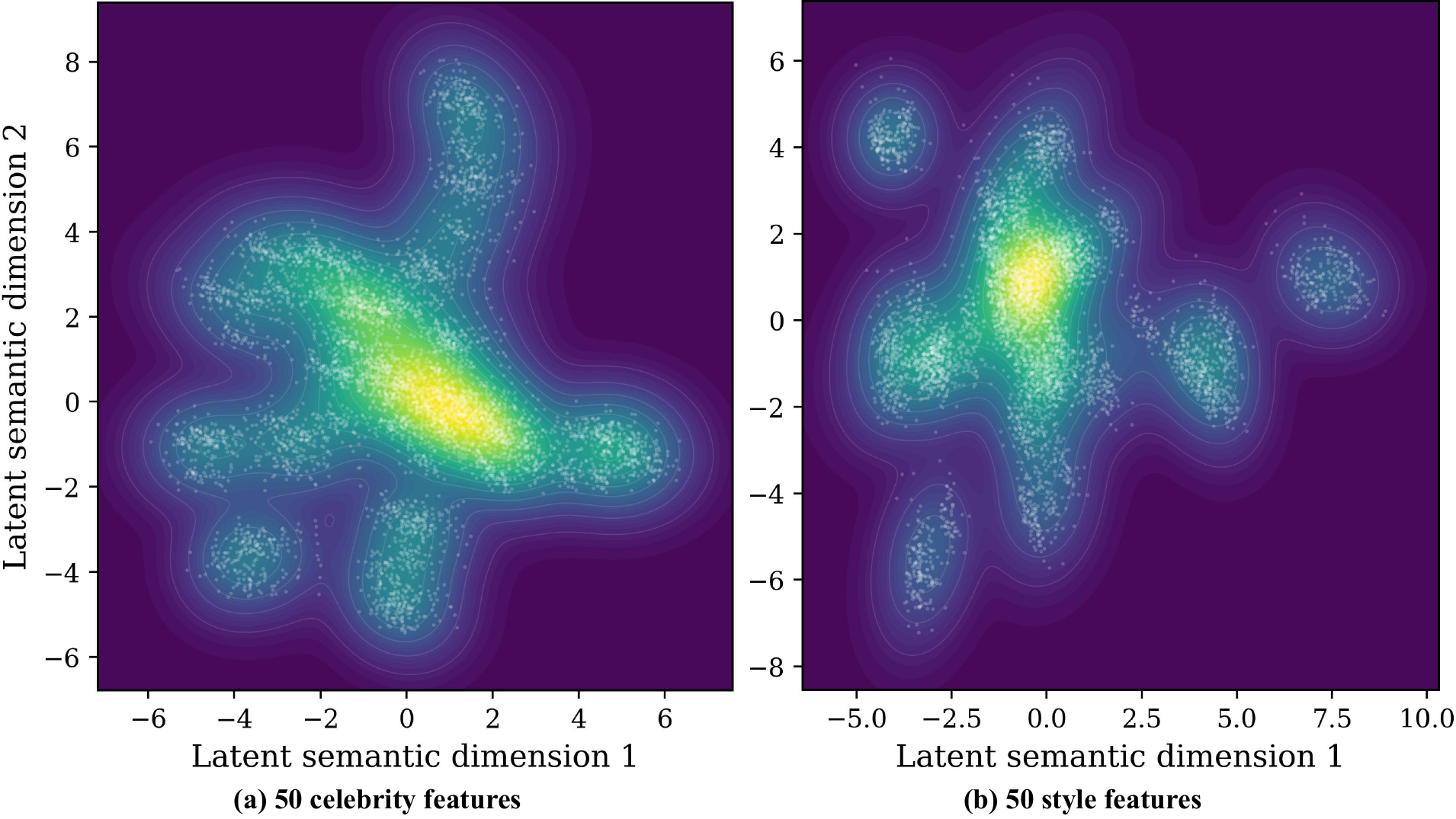}
\caption{
Semantic density of 50 celebrity targets and 50 artistic-style targets in the CLIP prompt embedding space.
Each point corresponds to one prompt, and the background shows the fitted GMM density after projection to two latent semantic dimensions.
}
\label{fig:50_feature}
\end{figure}

To understand why a single Gaussian Core LoRA can handle large-scale identity or style erasure, we visualize the feature distribution of 50 target concepts in the CLIP prompt embedding space. We extract prompt-level embeddings, project them into two latent semantic dimensions, and fit a GMM to estimate the target density, as shown in Figure~\ref{fig:50_feature}.

The visualization shows that the target concepts are not completely separated in the feature space. Many prompts overlap and form several high-density regions, indicating that large-scale identities or styles share common semantic structures. This supports the use of a shared low-rank erasure space, rather than assigning an independent LoRA adapter or isolated rank block to each target.

Meanwhile, the features do not collapse into a single compact cluster. Multiple local modes still exist, suggesting that different identities or styles may require different local erasure directions. A fixed LoRA update would average these modes and may lead to incomplete erasure or unnecessary disturbance to retained concepts. Gaussian Core LoRA addresses this by using GMM posterior responsibilities to dynamically reconfigure the shared LoRA rank space for each prompt. Therefore, Figure~\ref{fig:50_feature} empirically supports our design: large-scale targets contain shared structures that enable a single adapter, but their local mode diversity still requires distribution-aware dynamic adaptation.

\subsection{Additional Results on 50-Style Erasure}

We further evaluate Gaussian Core LoRA in a large-scale multi-style erasure setting, where the model is required to erase up to 50 target artistic styles using a single adapter. Compared with unsafe concepts or identity concepts, artistic styles are mainly reflected in relatively concentrated visual appearance patterns, such as color distribution, brushstroke texture, composition preference, and local rendering statistics. This makes multi-style erasure a suitable setting for evaluating whether a shared low-rank erasure space can capture common style-removal structures. Meanwhile, different artists and style families still correspond to distinct visual modes, which makes a purely static LoRA update prone to mode averaging.

\begin{figure*}[t]
\centering
\includegraphics[width=0.95\textwidth]{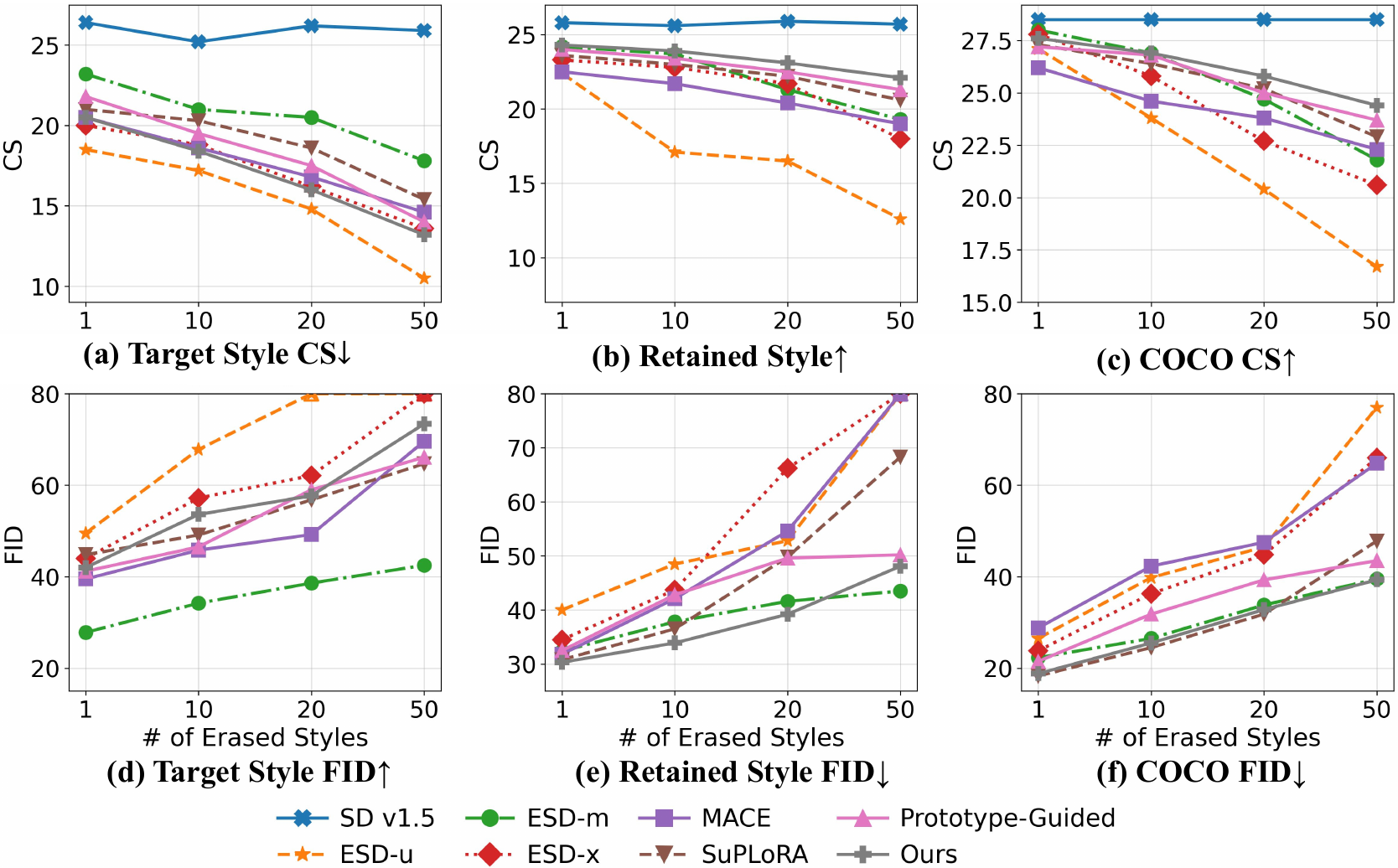}
\caption{
Scalability of multi-style erasure from 1 to 50 target styles.
We evaluate target-style removal by Target Style CS and Target Style FID, where lower Target Style CS and higher Target Style FID indicate stronger erasure.
Preservation is measured on retained styles and COCO using CLIP Score and FID, where higher CS and lower FID indicate better visual fidelity and semantic consistency.
Ours maintains stronger target-style suppression while introducing less distributional drift on retained styles and COCO as the number of erased styles increases.
}
\label{fig:style_scalability}
\end{figure*}

Figure~\ref{fig:style_scalability} reports the quantitative scalability results from 1 to 50 erased styles. We evaluate target-style removal using Target Style CLIP Score and Target Style FID, where lower Target Style CLIP Score and higher Target Style FID indicate stronger suppression of the erased styles. We evaluate preservation on retained styles and COCO prompts using CLIP Score and FID. As the number of target styles increases, static LoRA-based baselines tend to either leave visible target-style evidence or introduce larger distributional drift on retained styles and general COCO generations. In contrast, Gaussian Core LoRA maintains stronger target-style suppression while better preserving retained-style and COCO generation quality. This shows that the proposed dynamic core does not simply increase the overall erasure strength, but adaptively reconfigures the shared LoRA rank space according to the latent style mode of each prompt.

\begin{figure*}[t]
\centering
\includegraphics[width=0.95\textwidth]{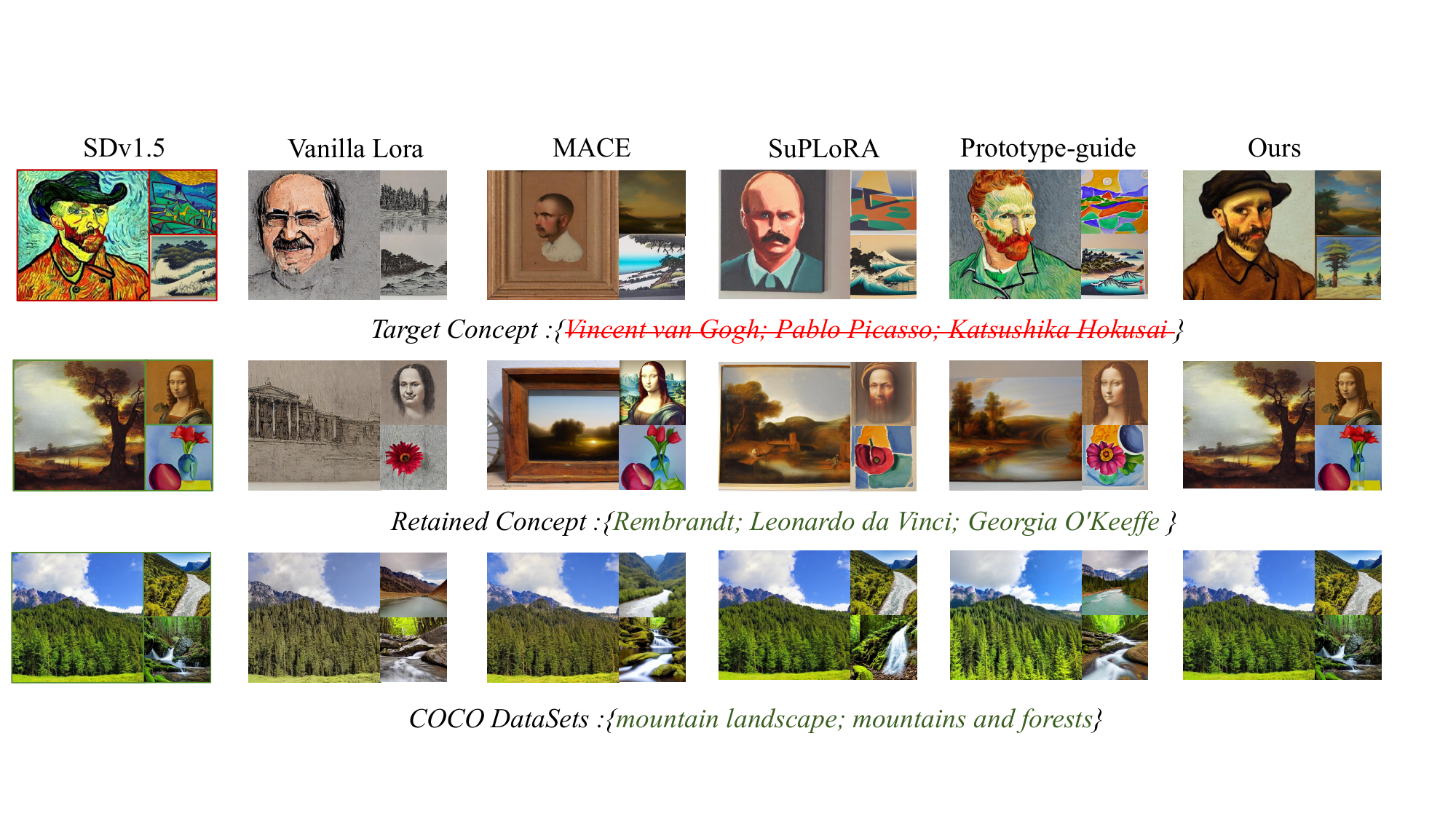}
\caption{
Qualitative comparison under the 50-style erasure setting.
The target styles include Vincent van Gogh, Pablo Picasso, and Katsushika Hokusai, while the retained styles include Rembrandt, Leonardo da Vinci, and Georgia O'Keeffe.
Gaussian Core LoRA suppresses the erased styles while preserving retained styles and general COCO scene semantics.
}
\label{fig:50_style}
\end{figure*}

Figure~\ref{fig:50_style} provides qualitative comparisons under the 50-style erasure setting. The target-style prompts include representative erased styles such as Van Gogh, Picasso, and Hokusai, while the retained prompts include non-target styles such as Rembrandt, Leonardo da Vinci, and Georgia O'Keeffe, together with general COCO scene prompts. The original model and static LoRA variants often retain recognizable target-style patterns or distort the retained concepts after strong erasure. By contrast, Gaussian Core LoRA effectively weakens the target artistic styles while preserving the visual identity of retained styles and the semantic structure of COCO scenes. These visual results are consistent with the quantitative trends and further demonstrate that broad style erasure can benefit from distribution-aware dynamic adaptation within a single shared adapter.

\subsection{Adaptation to FLUX}
\label{app:flux_adaptation}

FLUX replaces the UNet denoiser of Stable Diffusion with a
Transformer backbone containing multimodal and single-stream
Transformer blocks. Gaussian Core LoRA retains the same
distribution-aware gate--router--core formulation and only
changes the backbone-specific injection locations.

Specifically, we insert Gaussian Core LoRA into the joint-attention
modules of all multimodal Transformer blocks, where text-condition
tokens and visual latent tokens are explicitly coupled. The subsequent
single-stream Transformer blocks remain frozen and unmodified. This
design localizes concept erasure to cross-modal semantic interactions
and avoids unnecessary perturbations to the later visual generation
process.

For each selected attention projection \(W_j\), the effective weight is
defined as
\begin{equation}
    W_j^{\prime}(z)
    =
    W_j + g(z) B_j C_j(z) A_j,
\end{equation}
where \(W_j\) is frozen, \(A_j\) and \(B_j\) are the shared LoRA bases
of the corresponding projection, and \(C_j(z)\) is its
prompt-conditioned dynamic core. Each projection is assigned an
individual learnable embedding \(e_j\), and its core is generated from
the concatenation of the GMM responsibility vector and the projection
embedding:
\begin{equation}
    C_j(z)
    =
    \mathcal{C}_{\phi}\big([\rho(z);e_j]\big).
\end{equation}

Within each multimodal Transformer block, the adapters are applied to
both the visual-stream attention projections
\begin{equation}
    \{
    \texttt{to\_q},
    \texttt{to\_k},
    \texttt{to\_v},
    \texttt{to\_out.0}
    \},
\end{equation}
and the corresponding context-stream projections
\begin{equation}
    \{
    \texttt{add\_q\_proj},
    \texttt{add\_k\_proj},
    \texttt{add\_v\_proj},
    \\
    \texttt{to\_add\_out}
    \}.
\end{equation}
The former reconfigure the update applied to visual latent tokens,
whereas the latter modulate the text-context representations involved
in joint attention. Applying the same gate--router--core mechanism to
both streams enables prompt-dependent control over their semantic
interaction without modifying the remaining Transformer components.

For FLUX, the target GMM and component-wise support radii
\(\{\tau_k\}_{k=1}^{K}\) are estimated separately for each erasure
task and remain fixed throughout adapter training and inference.
The support percentile \(\beta\) and gate temperature \(T_g\) follow
the task-specific settings described in Appendix~\ref{app:details}. 

Because FLUX adopts a flow-matching parameterization rather than the
noise-prediction parameterization used by Stable Diffusion v1.5, we
replace the denoising prediction in the paired alignment objective with
the native FLUX flow prediction. Given a target--safe prompt pair
\((p_i^-,p_i^+)\), the FLUX pair loss is
\begin{equation}
\mathcal{L}_{\mathrm{pair}}^{\mathrm{FLUX}}
=
\mathbb{E}_{(p_i^-,p_i^+),t}
\left[
\left\|
f_{\theta+\Delta\psi}(x_t,t,p_i^-)
-
f_{\theta}(x_t,t,p_i^+)
\right\|_2^2
\right],
\end{equation}
where \(f_{\theta+\Delta\psi}\) denotes the FLUX Transformer equipped
with Gaussian Core LoRA and \(f_{\theta}\) is the frozen original
model. The same noisy latent \(x_t\) and timestep \(t\) are used for
the two prediction branches. The corresponding retention objective is
\begin{equation}
\mathcal{L}_{\mathrm{retain}}^{\mathrm{FLUX}}
=
\mathbb{E}_{p_i^+,t}
\left[
\left\|
f_{\theta+\Delta\psi}(x_t,t,p_i^+)
-
f_{\theta}(x_t,t,p_i^+)
\right\|_2^2
\right].
\end{equation}
The final FLUX training objective follows
\begin{equation}
    \mathcal{L}^{\mathrm{FLUX}}
    =
    \mathcal{L}_{\mathrm{pair}}^{\mathrm{FLUX}}
    +
    \lambda_r
    \mathcal{L}_{\mathrm{retain}}^{\mathrm{FLUX}}.
\end{equation}

Therefore, adapting Gaussian Core LoRA to FLUX requires neither a new
routing mechanism nor an architecture-specific core generator. Only
the adapter injection locations and the backbone-native prediction
parameterization are changed, while the Gaussian support gate, GMM-based soft routing, bounded dynamic core, and joint optimization objective remain unchanged.

\subsection{Cross-Architecture Qualitative Results on Sexual Erasure}

We further provide qualitative results on SDXL and FLUX to evaluate whether Gaussian Core LoRA can generalize beyond Stable Diffusion v1.5. These two models differ from SD v1.5 in model scale, denoising architecture, and generation behavior, making them suitable for examining the architecture compatibility of our method.

\begin{figure*}[t]
\centering
\includegraphics[width=0.8\textwidth]{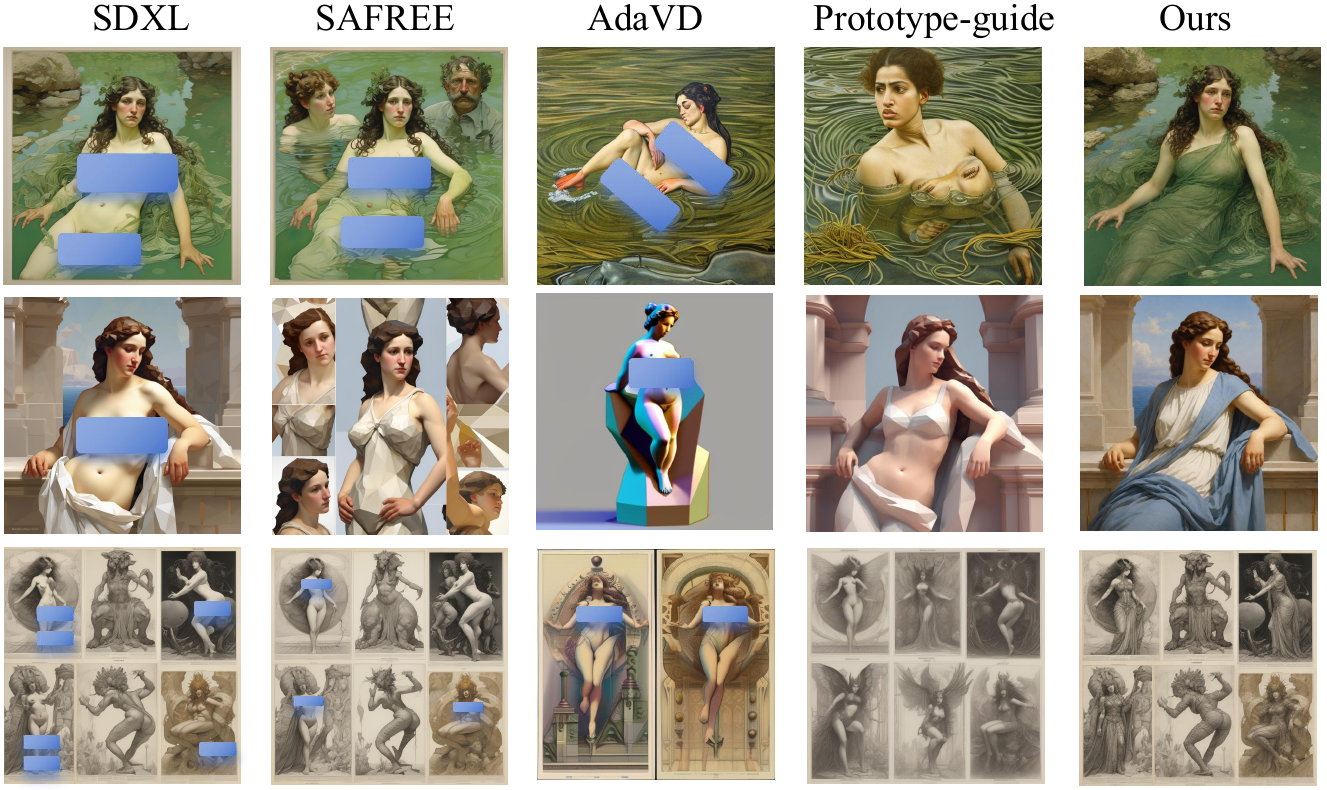}
\caption{
Qualitative results of Sexual erasure on SDXL.
Gaussian Core LoRA effectively suppresses Sexual-related visual cues while preserving the non-target semantics and overall scene structure of the input prompts, showing less semantic drift than baseline methods.
}
\label{fig:sdxl}
\end{figure*}

\begin{figure*}[t]
\centering
\includegraphics[width=0.8\textwidth]{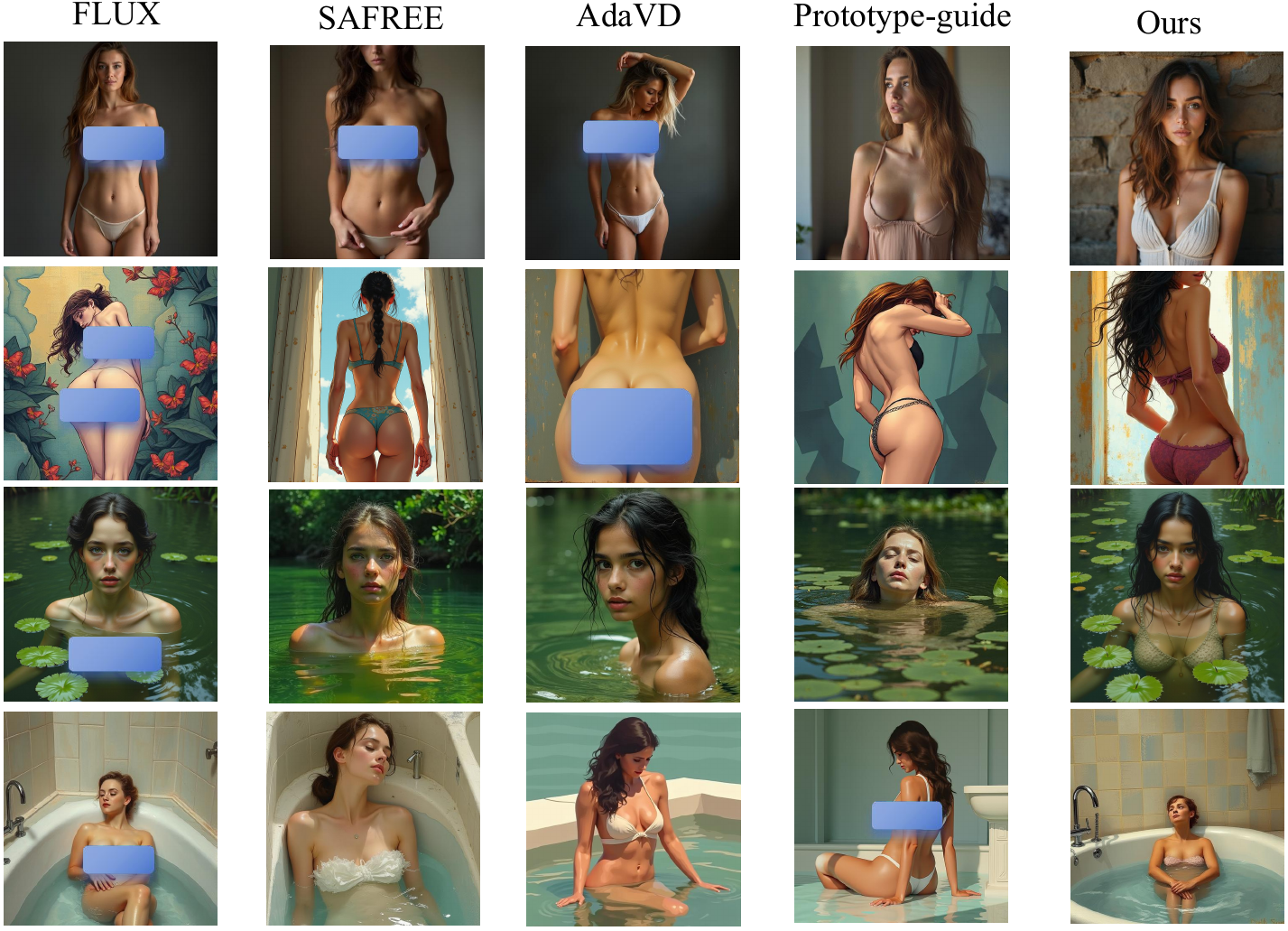}
\caption{
Qualitative results of Sexual erasure on FLUX.
Our method maintains target-concept suppression and better preserves prompt-relevant visual content, demonstrating the cross-architecture applicability of the proposed dynamic rank-space reconfiguration.
}
\label{fig:flux}
\end{figure*}

As shown in Figure~\ref{fig:sdxl} and Figure~\ref{fig:flux}, Gaussian Core LoRA achieves effective Sexual erasure on both SDXL and FLUX. Compared with baseline methods, our method more consistently suppresses the target Sexual-related visual cues while preserving the main non-target semantics of the prompt, such as scene layout, object context, and overall image structure. This indicates that the proposed dynamic core does not simply over-suppress the generation process, but performs a more localized intervention in the shared LoRA rank space.

These qualitative results are consistent with the quantitative cross-architecture results in the main paper. They suggest that distribution-aware dynamic adaptation is not tied to a specific diffusion backbone, and can be applied to different text-to-image models while maintaining a favorable balance between erasure effectiveness and semantic preservation.

\end{document}